\documentclass{article}

\usepackage{microtype}
\usepackage{graphicx}
\usepackage{caption}
\usepackage{subcaption}
\usepackage{booktabs} 

\usepackage{hyperref}

\usepackage[accepted]{icml2023}

\usepackage{amsmath}
\usepackage{amssymb}
\usepackage{mathtools}
\usepackage{amsthm}

\usepackage[capitalize,noabbrev]{cleveref}

\theoremstyle{plain}
\newtheorem{theorem}{Theorem}[section]
\newtheorem{proposition}[theorem]{Proposition}
\newtheorem{lemma}[theorem]{Lemma}

\theoremstyle{definition}
\newtheorem{definition}[theorem]{Definition}

\theoremstyle{remark}
\newtheorem{remark}[theorem]{Remark}

\usepackage[textsize=tiny]{todonotes}

\def\indep{\perp\!\!\!\perp}

\newcommand{\edited}[1]{#1}

\newcommand\footnoteref[1]{\protected@xdef\@thefnmark{\ref{#1}}\@footnotemark}

\icmltitlerunning{Counterfactual Identifiability of Bijective Causal Models}

\begin{document}

\twocolumn[
\icmltitle{Counterfactual Identifiability of Bijective Causal Models} 



\icmlsetsymbol{equal}{*}

\begin{icmlauthorlist}
\icmlauthor{Arash Nasr-Esfahany}{eecs}
\icmlauthor{Mohammad Alizadeh}{eecs}
\icmlauthor{Devavrat Shah}{eecs}
\end{icmlauthorlist}

\icmlaffiliation{eecs}{Department of Electrical Engineering and Computer Science (EECS), Massachusetts Institute of Technology (MIT), Cambridge, MA 02139}

\icmlcorrespondingauthor{Arash Nasr-Esfahany}{arashne@mit.edu}

\icmlkeywords{Machine Learning, Generation Mechanism, Counterfactual, Causal, Generative Model, Graphical Model, Identifiability, Computer Systems, Video Streaming, Unbiased Simulation, ICML}

\vskip 0.3in
]



\printAffiliationsAndNotice{}  

\begin{abstract}
We study counterfactual identifiability in causal models with bijective generation mechanisms (BGM), a class that generalizes several widely-used causal models in the literature. We establish their counterfactual identifiability for three common causal structures with unobserved confounding, and propose a practical learning method that casts learning a BGM as structured generative modeling. Learned BGMs enable efficient counterfactual estimation and can be obtained using a variety of deep conditional generative models. We evaluate our techniques in a visual task and demonstrate its application in a real-world video streaming simulation task.
\end{abstract}
\section{Introduction}
\label{sec:intro}

\stepcounter{footnote}
\emph{Had Cleopatra's nose been shorter, the whole face of \\the world would have changed.\footnote{Taken from \citet[Ch.~8]{pearl2018book}} (Blaise Pascal, 1669)}
\smallskip

The \emph{ladder of causation} presented by Pearl~\yrcite{pearl2018book} consists of three distinct layers encoding different types of concepts: \emph{associational}~($\mathcal{L}_1$), \emph{interventional}~($\mathcal{L}_2$), and \emph{counterfactual}~($\mathcal{L}_3$), roughly corresponding to \emph{seeing}, \emph{doing}, and \emph{imagining}, respectively.
$\mathcal{L}_1$ deals with passively observed factual information, for instance, the probability of recovery in patients who take Aspirin.
$\mathcal{L}_2$ deals with active interventions or the effect of actions, for example, what percentage of patients would recover if we give them Aspirin?
$\mathcal{L}_3$ deals with alternative ways the world could have been including ways that might conflict with how the world currently is, e.g., if the patient took Aspirin and was cured, would the headache still be gone had they not taken Aspirin?
The three levels are distinct in the sense that it is generally not possible to uniquely answer higher level queries from lower level information~\citep[Thm.~27]{bareinboim2022pearl}.
In particular, although we can determine $\mathcal{L}_2$ information by conducting experiments and actively intervening in the world, we cannot answer $\mathcal{L}_3$ questions in general, even with experiments.
In other words, they may be \emph{non-identifiable}~\cite{ibeling2020probabilistic}, or underspecified~\cite{d2020underspecification}.

Nevertheless, counterfactual queries of the form ``\emph{why--?}" and ``\emph{what if--?}" are useful in defining fundamental concepts such as harm~\cite{2022counterfactual-harm}, fairness~\cite{2017counterfactual-fairness,zhang2018fairness}, credit~\cite{2021counterfactual-credit-assignment}, regret, and blame.
They also have applications in engineering, e.g., root cause analysis~\cite{2022sage,2022causal-rootcause}, trace-driven simulation~\cite{causalsim,bothra2022veritas}, and sample efficient reinforcement learning~\cite{lu2020sample,agarwal2021persim}.

In this paper, we introduce a new class of causal models called \emph{Bijective Generation Mechanisms (BGM)}.
BGMs subsume several model classes studied in the literature, e.g., Nonlinear Additive Noise Models (ANM)~\cite{hoyer2008nonlinear}, Location Scale Noise Models (LSNM)~\cite{immer2022identifiability}, and post-nonlinear causal models (PNL)~\cite{zhang2009identifiability} (\S\ref{sec:problem}).
%
We establish counterfactual identifiability of BGMs for three meaningful causal structures
(\S\ref{sec:identifiability}), assuming a coarse knowledge of the underlying system in the form of a causal diagram.

Our identifiability results specify tractable criteria for finding the underlying BGM (up to equivalence, see \Cref{def:g-equivalence}), enabling counterfactual estimation.
We cast the search for a BGM that satisfies the specified criteria as density estimation with \emph{structured generative networks} (\S\ref{sec:learning}), which has been widely studied in the
learning literature~\cite{kingma2013auto,goodfellow2014generative,dinh2016density,arjovsky2017wasserstein,song2019generative,ho2020denoising}. 

Once the underlying BGM is learned, we use it for efficient counterfactual estimation with guarantees (\S\ref{sec:counterfactual}).
This is in contrast to symbolic counterfactual identification methods ~\cite{shpitser2007counterfactuals,correa2021nested} which express identifiable counterfactual queries in terms of interventional and observational distributions, statements that may not be computationally tractable. Furthermore, our work advances the growing literature on applying machine learning methods for counterfactual estimation. Unlike most prior work in this space that foregoes identifiability analysis~\cite{pawlowski2020deep,sanchez2021diffusion}, except in certain restricted settings (e.g., discrete domains~\cite{shalit2017estimating,oberst2019counterfactual} or absence of unobserved confounders~\cite{khemakhem2021causal,sanchez2021vaca}), our results support continuous variables, allow unobserved confounders, and extend to a more general class of generation mechanisms. 


We validate our techniques using a visual task (\S\ref{eval:ellipse}) and demonstrate their application to a real-world system simulation task (\S\ref{subsec:trace-driven}).
Our code is available in \href{github.com/counterfactual-BGM/cf2}{github.com/counterfactual-BGM/cf2}. This work does not raise any negative societal impacts.

\section{Preliminaries}
\label{sec:prelim}

We provide a brief background on the (mostly) causal concepts that we use in this work.
Refer to \citet{pearl2009causality}, \citet{pearl2009primer}, and \citet{peters2017elements} for more details.

\textbf{Notation:}
We refer to a random variable with a capital letter, e.g., $X$, the value it obtains with a lowercase letter, e.g., $x$, its Probability Density Function (PDF) with $P_X$, and a set of random variables with boldface font, e.g., $\boldsymbol{X} = \{X_1, \ldots, X_n\}$.
$J_{f(\cdot)}$ denotes the Jacobian of $f(\cdot)$.

\textbf{SCM and Causal Diagram:}
We use the framework of Structural Causal Models (SCMs)~\citep[Ch.~7]{pearl2009causality}.
An SCM $\mathcal{M}$ consists of endogenous variables $\boldsymbol{V}$ determined by other variables in the model, exogenous (also called latent or background) variables $\boldsymbol{U}$ with distribution $P_{\boldsymbol{U}}(\cdot)$ determined by factors outside the model (one exogenous variable corresponding to each endogenous variable), and generation mechanisms $\mathcal{F}$.
$\mathcal{F}$ contains a function $f_i$ for each variable $V_i$ that maps endogenous parents $\text{Pa}(V_i)$ and exogenous variable $U_i$ to $V_i$.
The entire $\mathcal{F}$ defines a structured mapping from $\boldsymbol{U}$ to $\boldsymbol{V}$.
The prior distribution over exogenous variables, $P_{\boldsymbol{U}}(\cdot)$, together with the generation mechanisms $(\mathcal{F})$ entail an observational $(\mathcal{L}_1)$ distribution over endogenous variables which we refer to as $P_{\boldsymbol{V}}^\mathcal{M}(\cdot)$.
Each $\mathcal{M}$ induces a causal diagram $\mathcal{G}$, where every $V_i \in \boldsymbol{V}$ is a vertex, there is a directed arrow $(V_j \rightarrow V_i)$ for every $V_i \in \boldsymbol{V}$ and $V_j \in \text{Pa}(V_i)$, and there is a dashed-bidirected arrow $(V_j \dashleftarrow \dashrightarrow V_i)$ for every $V_i, V_j \in \boldsymbol{V}$ such that $U_i$ and $U_j$ are not independent.
The bidirected arrows represent existence of unobserved confounders.
A causal model satisfies the \emph{Markovian} assumption if for every $V_i, V_j \in \boldsymbol{V}$, the corresponding $U_i$ and $U_j$ are independent.
In other words, no bidirected arrow exists in its induced causal diagram $(\mathcal{G})$.

\textbf{Interventions and the \textit{do}-Operator}: 
Given an SCM $\mathcal{M}$, and a set of endogenous variables $\boldsymbol{X} \subseteq \boldsymbol{V}$, an intervention $\textit{do}(\boldsymbol{X} \coloneqq \boldsymbol{x})$ corresponds to replacing the generation mechanisms $\mathcal{F}$ with $\mathcal{F}_{\boldsymbol{x}} = \{f_i : V_i \notin \boldsymbol{X}\} \cup \{\boldsymbol{X} \coloneqq \boldsymbol{x}\}$.
In words, $\mathcal{F}_{\boldsymbol{x}}$ is formed by deleting from $\mathcal{F}$ all functions $f_i$ corresponding to members of set $\boldsymbol{X}$ and replacing them with the set of constant functions $\{\boldsymbol{X} \coloneqq \boldsymbol{x}\}$.
We refer to the altered SCM by $\mathcal{M}_{\boldsymbol{x}}$, and the interventional $(\mathcal{L}_2)$ distribution of endogenous variables by $P_{\boldsymbol{V}}^{\mathcal{M}_{\boldsymbol{x}}}(\cdot)$ or in short $P_{\boldsymbol{V}_{\boldsymbol{x}}}(\cdot)$.
The interventional (or experimental) distribution helps us analyze the effect of taking actions, i.e., what would happen if we apply an intervention?

\textbf{Counterfactuals}: 
Given an SCM $\mathcal{M}$, two sets of endogenous variables $\boldsymbol{X}, \boldsymbol{E} \subseteq \boldsymbol{V}$, observed realizations $\boldsymbol{e}$ (evidence) for $\boldsymbol{E}$, and an intervention $\textit{do}(\boldsymbol{X} := \boldsymbol{x})$, the counterfactual $(\mathcal{L}_3)$ distribution $P_{\boldsymbol{V}}^{\mathcal{M}_{\boldsymbol{x}}}(\cdot|\boldsymbol{E} = \boldsymbol{e})$ or in short $P_{\boldsymbol{V}_{\boldsymbol{x}}}(\cdot|\boldsymbol{E} = \boldsymbol{e})$ corresponds to the distribution entailed by $\mathcal{M}_{\boldsymbol{x}}$ using the posterior distribution $P_{\boldsymbol{U}|\boldsymbol{E}}(\cdot|\boldsymbol{e})$ over the exogenous variables.
In case of deterministic counterfactuals ($\delta$-distribution), we refer to them as $\boldsymbol{V}_{\boldsymbol{x}}|\boldsymbol{E} = \boldsymbol{e}$.
\citet[Ch.~7]{pearl2009causality} proposes the following three-step procedure for counterfactual estimation.
i) \textit{Abduction:} Update $P_{\boldsymbol{U}}$ with $\boldsymbol{e}$ to obtain $P_{\boldsymbol{U}|\boldsymbol{E}}$.
ii) \textit{Action:} Update the SCM $\mathcal{M}$ to $\mathcal{M}_{\boldsymbol{x}}$.
iii) \textit{Prediction:} Use the updated distribution of exogenous variables and the updated SCM to estimate the counterfacual distribution.
For generalizations of this definition, e.g., to nested counterfactuals, refer to \citet{correa2021nested}.
Counterfactual distributions allow us to imagine hypothetical worlds where everything is fixed other than interventions. 

\textbf{Identifiability}:
Evaluating causal queries given a partial state of knowledge is a subtle problem. An $\mathcal{L}_i$ query is identifiable using $\mathcal{L}_j$ information if its answer can be expressed purely based on $\mathcal{L}_j$ distributions $(i,j \in \{1,2,3\})$.
Formally, let $Q(\mathcal{M})$ be an $\mathcal{L}_i$ query of an SCM $\mathcal{M}$.
In a class $\mathbb{M}$ of SCMs, $Q$ is identifiable if for any pair of SCMs $\mathcal{M}_1$ and $\mathcal{M}_2$ from $\mathbb{M}$, $Q(\mathcal{M}_1) = Q(\mathcal{M}_2)$ whenever $\mathcal{M}_1$ and $\mathcal{M}_2$ match in all $\mathcal{L}_j$ queries~\citep[def.~3.2.3]{pearl2009causality}. 
This is always true if $1 \leq i \leq j \leq 3$.
Characterizing conditions where this holds for $1\leq j < i \leq 3$ has been subject to extensive research efforts~\cite{spirtes2001causation}.
\section{Related Work}
\label{sec:related}

\textbf{Interventional $(\mathcal{L}_2)$ Identification and Estimation}:
Assuming a coarse knowledge of the underlying system in the form of a causal DAG, identification of interventional queries has been extensively studied in the literature~\citep[Ch.~3]{pearl2009causality}, including \textit{do-calculus}~\cite{pearl1995causal} as a general solution. 
In cases where the full causal diagram is not accessible, another line of work focuses on its (partial) identification using observational data.
Although this sounds compelling and amenable to data-driven and ML research practices, it is known that the causal diagram can be identified from observational data only up to its Markovian equivalence class~\cite{spirtes2001causation}.
Investigating identifiability using only equivalence classes has thus received research attention~\cite{zhang2008completeness,perkovic2018complete,jaber2019causal}.
\citet{witti2021sbi} investigates a simulation-based notion of identifiability with probabilistic programs~\cite{goodman2008church}, using priors over parametric representations of SCMs.
Once identifability of the interventional query is established, various methods exist for estimation of the causal effect, including Propensity Score for the backdoor case~\cite{rubin1978bayesian,kennedy2019nonparametric}, and other statistical methods for more relaxed settings~\cite{fulcher2020robust,jung2020learning}.
In cases where the interventional query is non-identifiable, \emph{partial identification} methods estimate meaningful bounds~\cite{manski1990nonparametric,balke1997bounds,zhang2021bounding,li2022bounds}

\textbf{Counterfactual $(\mathcal{L}_3)$ Identification}:
Using the \emph{counterfactual graph}, \citet{shpitser2007counterfactuals} proposes an algorithm that determines identifiability of counterfactual queries from interventional data.
This was extended in \citet{correa2021nested} by providing sufficient and necessary graphical conditions for identification of (nested) counterfactuals from a collection of observational and interventional distributions, given the causal diagram.
In identifiable cases, these algorithms express the counterfactual query as a combination of observational and interventional quantities.
However, estimating this expression may not be tractable.
\citet{shah2022counterfactual} proves identifiability of counterfactuals for a specific causal diagram and provides an algorithm for their tractable estimation, assuming the joint distribution of exogenous and endogenous variables is an exponential family.
In contrast, we do not restrict the joint distribution of variables.

Similar to the interventional case, partial identification methods have been proposed for non-identifiable counterfactual queries~\cite{balke1994counterfactual,tian2000probabilities,finkelstein2020deriving,zhang2022partial,gresele22}.
Furthermore, identification of specific counterfactual queries has been studied in isolation, e.g., the \emph{effect of treatment on the treated}~\cite{shpitser2009effects}, \emph{path-specific effects}~\cite{shpitser2018identification,zhang2018fairness}, and \emph{probabilities of causation}~\cite{pearl1999probabilities}.

\textbf{Neural Methods for Causal Estimation\footnote{\Cref{app:related} provides a more detailed survey.}}:
There is a growing literature on applying neural methods for efficient estimation of causal queries.
An extensive line of work focuses on estimating interventional $(\mathcal{L}_2)$ queries including \citet{kocaoglu2018causalgan,xia2021causal,zevcevic2021relating}, to name a few.
However, we focus on estimating counterfactuals $(\mathcal{L}_3)$.

On the counterfactual side, a line of work uses ML techniques for estimation without any guarantees about the identifiability of the counterfactual query~\cite{pawlowski2020deep,khemakhem2021causal,sanchez2021vaca,sanchez2021diffusion,geffner2022deep}.
In contrast, we prove identifiability for several causal structures in \S\ref{sec:identifiability}. Furthermore, most existing techniques make restrictive assumptions about  the causal structure or generation mechanisms. For example, \citet{johansson2016learning,shalit2017estimating,yao2018representation,oberst2019counterfactual,lorberbom2021learning,xia2022neural} consider counterfactual estimation in domains with discrete (finite-valued) endogenous variables. Several others works assume a Markovian causal structure~\cite{pawlowski2020deep,khemakhem2021causal,sanchez2021vaca,geffner2022deep}, i.e., the absence of unobserved confounders (bidirected edges in the causal diagram).
\citet{hartford2017deep,khemakhem2021causal,geffner2022deep} restrict the class of generation mechanisms to Nonlinear Additive Noise Models (ANM)~\cite{hoyer2008nonlinear} or Location Scale Noise Models (LSNM)~\cite{strobl2022identifying}. The models we consider are more general than these prior papers. They support continuous endogenous variables, allow unobserved confounders, and bijective generation mechanisms that include nonlinear ANM, LSNM, and post-nonlinear causal model (PNL)~\cite{zhang2009identifiability} as special cases.

\textbf{Disentanglement}:
Independent Component Analysis (ICA)~\cite{hyvarinen2000independent} concerns the problem of recovering statistically independent source signals $\boldsymbol{S} = (S_1, \ldots, S_n)$ from their observed mixtures $\boldsymbol{X} = (X_1, \ldots, X_n)$, where $\boldsymbol{X} = f(\boldsymbol{S})$, and the unknown $f$ (mixing function) is assumed to be invertible.
Unlike linear mixing functions that make this problem identifiable with certain conditions, it is known to be non-identifiable for non-linear mixing functions~\cite{hyvarinen1999nonlinear}.
Recently, \citet{hyvarinen2019nonlinear} proved identifiability in the presence of auxiliary variables, which make the sources conditionally independent.
This has been exploited for disentangling semantically meaningful features of high-dimensional data~\cite{locatello2019challenging,khemakhem2020variational}.

We took inspiration from the non-linear ICA framework, especially for our development in \S\ref{sub:BC}.
However, the problem we consider is fundamentally different: We are interested in disentangling the total contribution of the unknown sources of variation from known attributes, as opposed to disentangling the effect of individual unknown sources.
A recent work~\cite{shaham2022discovery} considers disentangling latent variables from observed attributes, and proves that the reconstructed sources $\hat{\boldsymbol{S}}$ have the same entropy as the true sources, in distribution.
However, we are interested in point-wise guarantees, i.e., $\hat{\boldsymbol{s}} = g(\boldsymbol{s})$ for some invertible function $g(\cdot)$, which are stronger than the distributional properties like $\mathcal{H}(\hat{\boldsymbol{S}}) = \mathcal{H}(\boldsymbol{S})$.
\section{Problem Formulation}
\label{sec:problem}
We assume knowledge of the causal diagram $\mathcal{G}$ that might include unobserved confounders (no Markovianity assumption), and access to observational ($\mathcal{L}_1$)
data \edited{distribution}.
We are interested in learning the data generation mechanisms $\mathcal{F}$, which can be further used for counterfactual $(\mathcal{L}_3)$ estimation.
As demonstrated by Pearl's Causal Hierarchy Theorem~\citep[Thm.~27]{bareinboim2022pearl}, cross-layer inference is not possible in general settings, and we need to make further assumptions.

\textbf{Bijective Generation Mechanism (BGM)}:
As mentioned in \S\ref{sec:prelim}, each endogenous variable $V_i$ is generated in the following way in the SCM framework:
\begin{equation}
\label{eqn:generation_mechanism}
    V_i \coloneqq f_i\Big(\text{Pa}(V_i), U_i\Big)
\end{equation}
We assume that the function $f_i$ is a bijective mapping from $U_i$ to $V_i$, for each realization of $\text{Pa}(V_i)$, hence the name \emph{Bijective Generation Mechanism (BGM)}.
In other words, no information is lost in transformation from exogenous to endogenous variables.
We refer to the inverse of the generation mechanism as $f_i^{-1}(\text{Pa}, \cdot)$, i.e.,
\begin{equation}
\label{eqn:reverse_generation_mechanism}
    U_i = f_i^{-1}\Big(\text{Pa}(V_i), V_i\Big).
\end{equation}
Note that the nonlinear ANM~\cite{peters2014causal}, LSNM~\cite{immer2022identifiability}, and PNL~\cite{zhang2009identifiability} models from prior work are all special cases of BGMs. Specifically, the nonlinear ANM model, 
\begin{equation}
    f_i\Big(\text{Pa}(V_i), U_i\Big) = g_i\Big(\text{Pa}(V_i)\Big) + U_i
\end{equation} LSNM~\cite{immer2022identifiability} model, 
\begin{equation}
    f_i\Big(\text{Pa}(V_i), U_i\Big) = l\Big(\text{Pa}(V_i)\Big) + s\Big(\text{Pa}(V_i)\Big)U_i
\end{equation}
with $s$ a strictly positive function on $\mathbb{R}$, 
and PNL causal models~\cite{zhang2009identifiability}, 
\begin{equation}
    f_i\Big(\text{Pa}(V_i), U_i\Big) = h_i\bigg(g_i\Big(\text{Pa}(V_i)\Big) + U_i\bigg)
\end{equation}
with $h_i$ an invertible function, are all bijective given $\text{Pa}(V_i)$.

\section{Identifiability}
\label{sec:identifiability}

\begin{figure}
    \centering
    \begin{subfigure}{0.49\linewidth}
        \centering
        \includegraphics[width=0.6\linewidth]{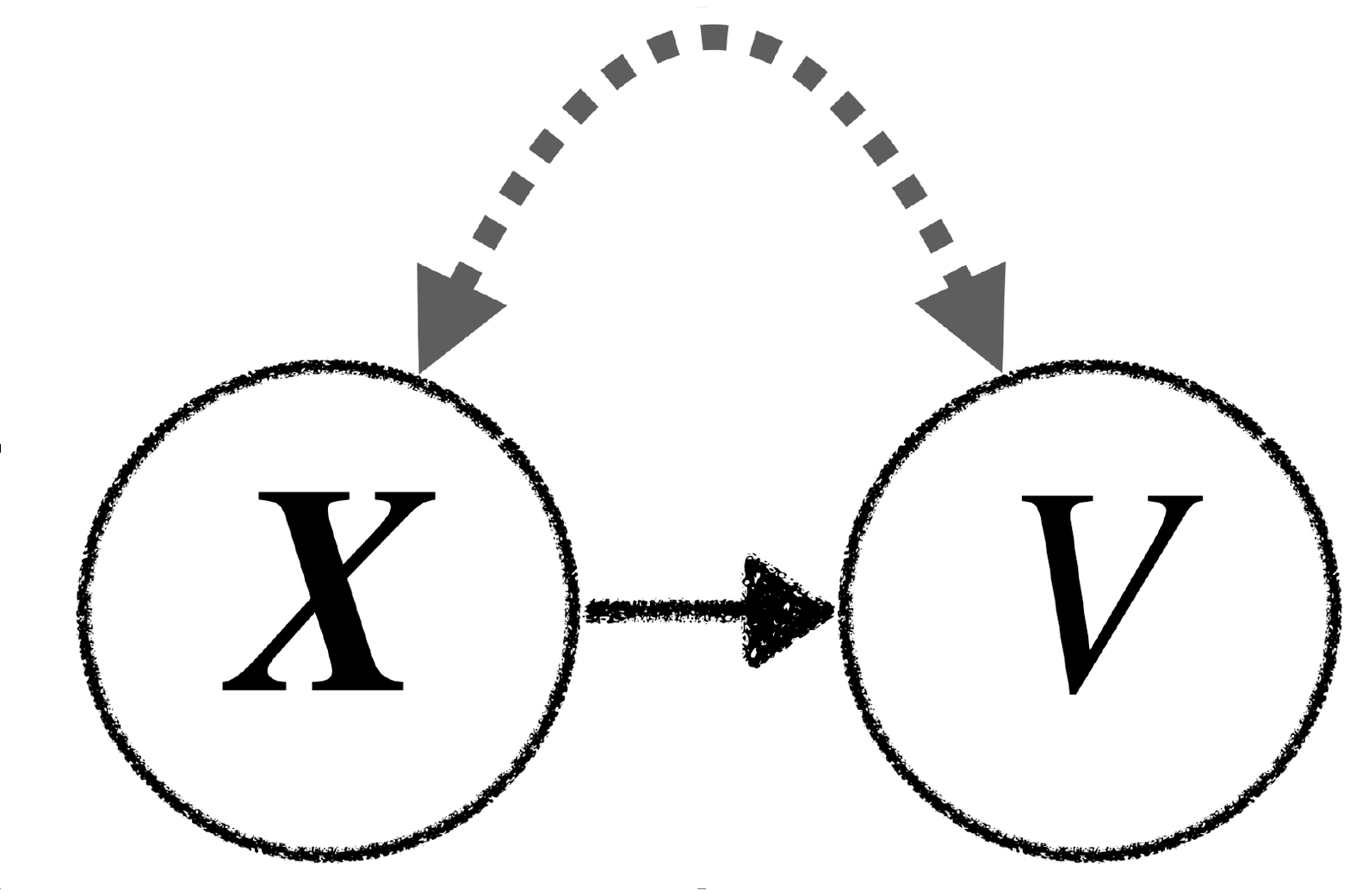}
        \caption{Non-Markovian}
        \label{subfig:sub-graph}
    \end{subfigure}
    \begin{subfigure}{0.49\linewidth}
        \centering
        \includegraphics[width=0.6\linewidth]{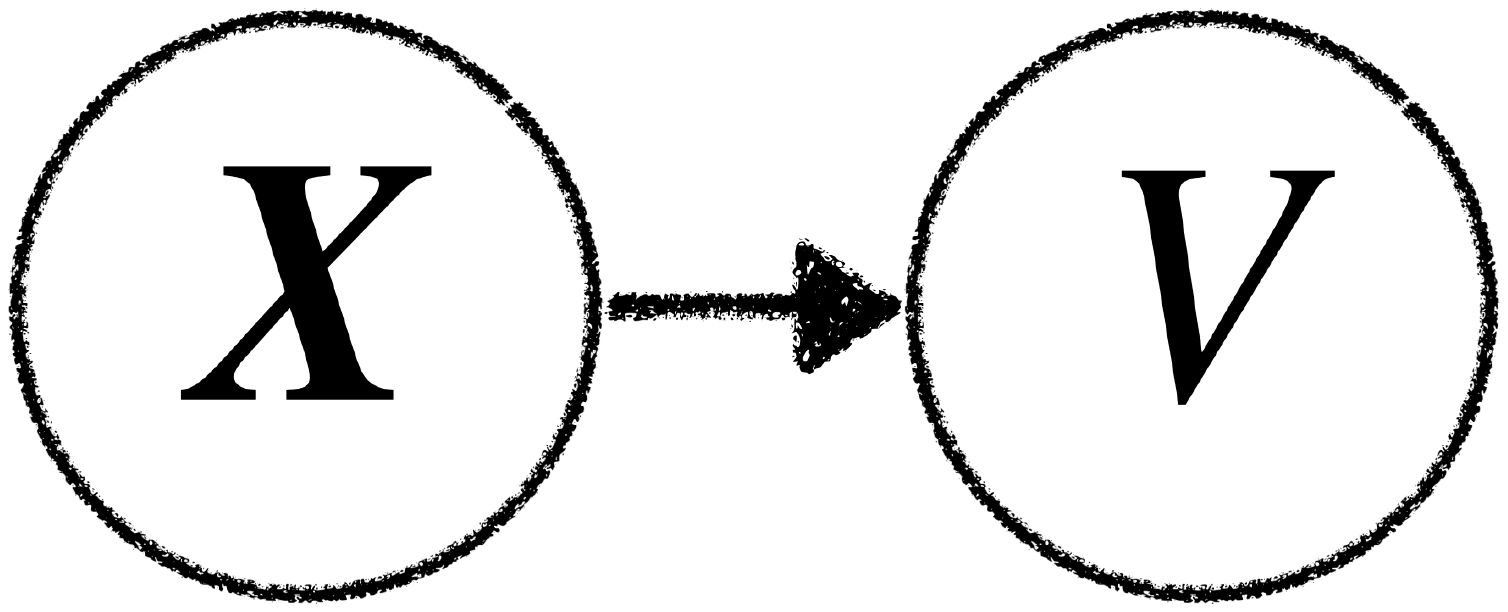}
        \caption{Markovian}
        \label{subfig:intervened}
    \end{subfigure}
    \caption{Causal diagram $\mathcal{G}$'s sub-graph related to generation of $V$}
    \label{fig:causal-dag}
\end{figure}

Without loss of generality, we focus on identifying the generation mechanism $f_i$ of a particular endogenous node $V_i$ shown in \Cref{eqn:generation_mechanism}.
For ease of exposition, we drop the subscript $i$, and refer to $\text{Pa}(V)$ as $\boldsymbol{X}$.
The sub-graph of $\mathcal{G}$ which is related to the generation of $V$ is depicted in \Cref{subfig:sub-graph}.
The dashed-bidirected arrow indicates potential existence of an unobserved confounder.
All proofs are in \Cref{app:proof}.


\edited{We provide three sets of constraints on $f$ and the underlying causal structure that imply counterfactual identifiability given an observed ($\mathcal{L}_1$) data distribution ($\mathcal{D}$), i.e., given $\mathcal{D}$, each set of constraints identifies $f$ up to indeterminacies that do not affect counterfactual queries.}

$U$ and $V$ can be single- or multi-dimensional \edited{in general}. Our results in \S\ref{sub:Markovian} and \S\ref{sub:IV} are for the scalar case, while the result in \S\ref{sub:BC} applies to multi-dimensional  $U$ and $V$.


\subsection{The Markovian Case}
\label{sub:Markovian}
If the exogenous variable $U$ associated with $V$ is \edited{independent of} its parents $\boldsymbol{X}$, we end up with the causal diagram shown in \Cref{subfig:intervened} where $\boldsymbol{X} \indep U$.\footnote{It is common practice to exclude the exogenous variable $U$ from the causal diagram.}


\edited{
\begin{theorem}
\label{thm:intervention}
BGM $f$ is counterfactually identifiable given $P_{\boldsymbol{X}, V}$ if
\begin{enumerate}
    \item (Markovian) $U \indep \boldsymbol{X}$.
    \item For all $\boldsymbol{x}$, $f(\boldsymbol{x}, \cdot)$ is either a strictly increasing function or a strictly decreasing function.
\end{enumerate}
\end{theorem}
}
This theorem is similar to \citet[Thm.~1]{lu2020sample}, and is mentioned here for completeness. In this result, \edited{$U, \boldsymbol{X}, V$} can be discrete or continuous. 
Note that 
independence of $U$ and $\boldsymbol{X}$ \edited{by itself} is not sufficient for \edited{identifiability} as demonstrated with a counter-example in \Cref{app:subsub:counter-example} and an experiment in \Cref{app:ellipse:failure}.
It is not clear how to generalize the monotonicity condition to BGMs with multi-dimensional $V$, which is a known issue in Markovain causal structures~\cite{non-ident}.

\subsection{Instrumental Variable (IV)}
\label{sub:IV}
\begin{figure}
    \centering
    \begin{subfigure}{0.49\linewidth}
        \centering
        \includegraphics[width=0.6\linewidth]{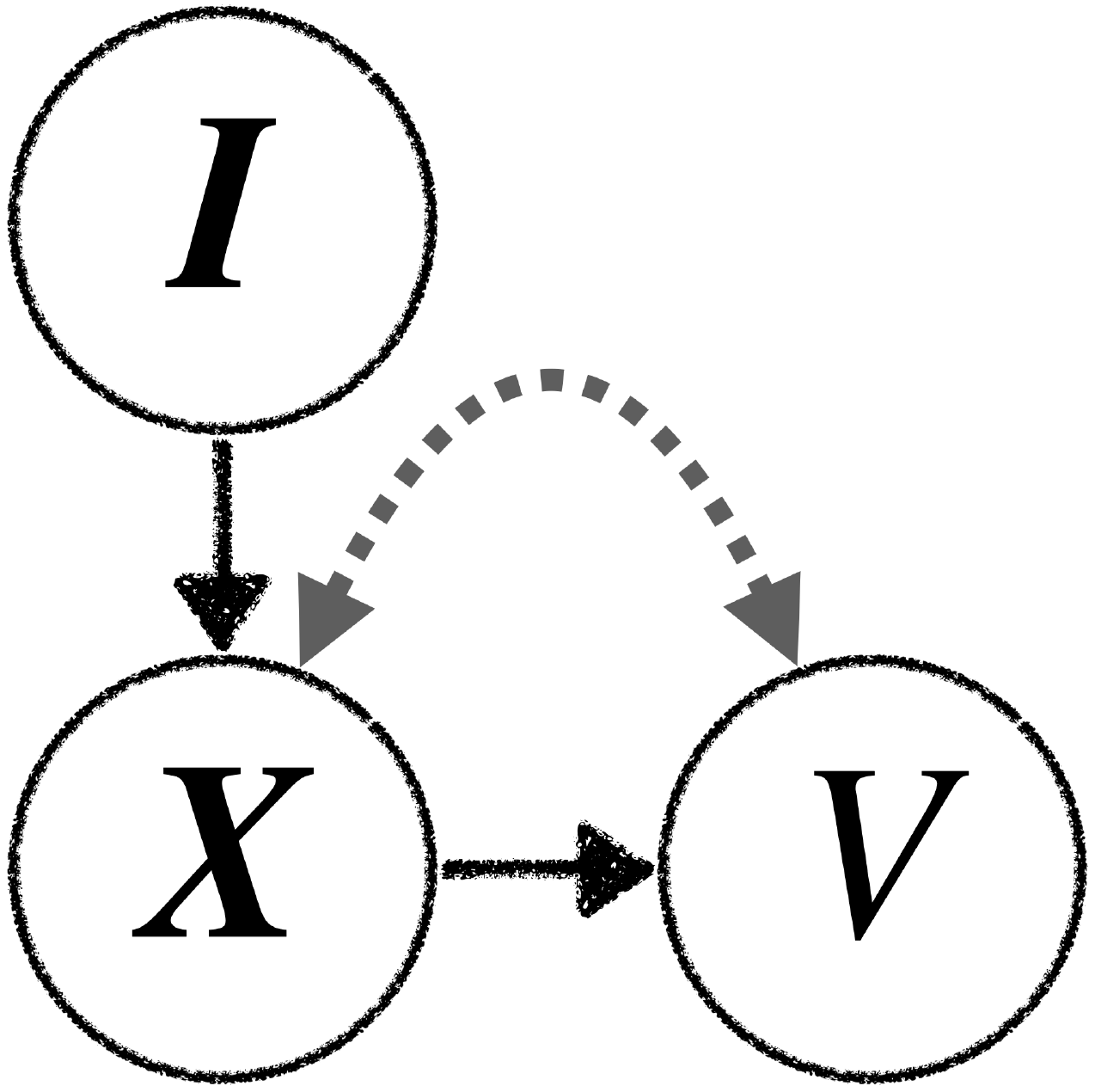}
        \caption{Instrumental Variable}
        \label{subfig:IV}
    \end{subfigure}
    \begin{subfigure}{0.49\linewidth}
        \centering
        \includegraphics[width=0.6\linewidth]{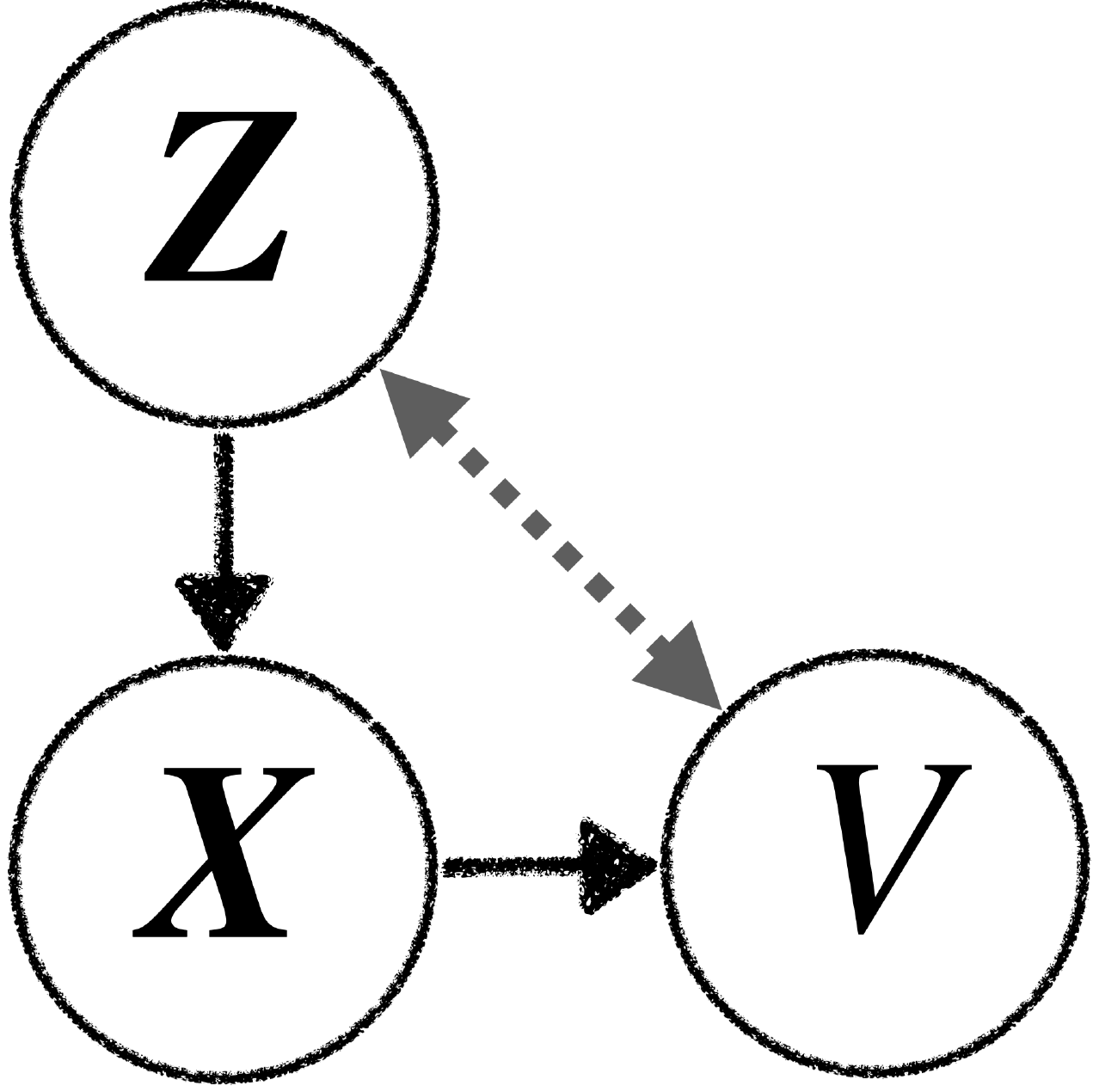}
        \caption{Backdoor Criterion}
        \label{subfig:BC}
    \end{subfigure}
    \caption{Instrumental Variable and Backdoor Criterion examples.}
\end{figure}
Even with an unobserved confounder, we can establish \edited{counterfactual identifiability} from observational $(\mathcal{L}_1)$ data if we can find IVs relative to the pair $(\boldsymbol{X}, V)$. We define an IV as a set of variables independent of $U$ that affect $V$ only through $\boldsymbol{X}$.
\Cref{subfig:IV} shows an example in which $\boldsymbol{I}$ is an IV with respect to $(\boldsymbol{X}, V)$.\footnote{There are several other IV definitions in the causal literature that capture the same concept, e.g., \citet{bowden1990instrumental,pearl1995testability,angrist1996identification,heckman1999local}.
See \citet[Sec.~7.4.5]{pearl2009causality} for a discussion.}

The following theorem formalizes \edited{counterfactual identifiability} in this setting for discrete-valued $\boldsymbol{X}$ and $\boldsymbol{I}$, i.e., $\boldsymbol{X} \in \mathbb{X} \triangleq \{\boldsymbol{x}_1, \ldots, \boldsymbol{x}_n\}$ and $\boldsymbol{I} \in \mathbb{I} \triangleq \{\boldsymbol{i}_1, \ldots, \boldsymbol{i}_n\}$.\footnote{If $|\mathbb{I}| > n$, pick any subset with $n$ members.}


\edited{
\begin{theorem}
\label{thm:IV}
BGM $f$ is counterfactually identifiable given $P_{\boldsymbol{X}, V, \boldsymbol{I}}$ if
\begin{enumerate}
    \item (IV) $\boldsymbol{I} \indep U$.
    \item For all $\boldsymbol{x} \in \mathbb{X}$, $f(\boldsymbol{x}, \cdot)$ and $f^{-1}(\boldsymbol{x}, \cdot)$ are either strictly increasing or strictly decreasing, and two times differentiable functions.
    \item $P(\boldsymbol{i}, \boldsymbol{x}, \cdot)$ is differentiable for every $\boldsymbol{i} \in \mathbb{I}, \boldsymbol{x} \in \mathbb{X}$.
    \item (Positivity) $\forall u, \boldsymbol{x} \in \mathbb{X}: P_{U, \boldsymbol{X}}(u, \boldsymbol{x}) > 0$.
    \item (Variability) $\forall u: |\det \boldsymbol{M}(u, \mathbb{I})| \ge c > 0$ , where 
    \begin{equation*}
    \boldsymbol{M}(u, \mathbb{I}) \triangleq    
    \begin{bsmallmatrix}
    P(\boldsymbol{x}_1|u,\boldsymbol{i}_1) & \ldots & P(\boldsymbol{x}_n|u,\boldsymbol{i}_1) \\
    \vdots                   & \ddots & \vdots                   \\
    P(\boldsymbol{x}_1|u,\boldsymbol{i}_n) & \ldots & P(\boldsymbol{x}_n|u,\boldsymbol{i}_n) \\
    \end{bsmallmatrix}
    \end{equation*}
\end{enumerate}
\end{theorem}
}

Besides the technical conditions required for the proof, the variability condition in \Cref{thm:IV} has the following interpretation: the IV must take a sufficient number of distinct values (at least as many values as possible for $\boldsymbol{X}$), and that the IV must have a strong influence on $\boldsymbol{X}$. Positivity implies that the support of U is independent of $\boldsymbol{X}$.

\subsection{The Backdoor Criterion (BC)}
\label{sub:BC}
The second setting in which we can establish counterfactual identifiability from observational $(\mathcal{L}_1)$ data in the presence of latent confounding is when there exists a set of variables $(\boldsymbol{Z})$ that satisfy the backdoor criterion (BC) with respect to $(\boldsymbol{X}, V)$, i.e., if $\boldsymbol{Z}$ blocks every path between $\boldsymbol{X}$ and $V$ that contains an arrow into $\boldsymbol{X}$.\footnote{\citet[Def.~3.3.1]{pearl2009causality} requires $\boldsymbol{Z}$ to be non-descendent of $\boldsymbol{X}$ as well, but we do not need such a requirement as our goal is not to use $\boldsymbol{Z}$ for adjustment.}
Intuitively, we want $\boldsymbol{Z}$ to be responsible for all the spurious correlation (the dashed-bidirected edge) between $\boldsymbol{X}$ and $U$.
\Cref{subfig:BC} demonstrates an example where $\boldsymbol{Z}$ satisfies the BC with respect to $(\boldsymbol{X}, V)$. 

In the following, assume $U, V \in \mathbb{R}^d$.

\edited{
\begin{theorem}
\label{thm:BC}
BGM $f$ is counterfactually identifiable given $P_{\boldsymbol{X}, V, \boldsymbol{Z}}$ if
\begin{enumerate}
    \item (BC) $U \indep \boldsymbol{X} | \boldsymbol{Z}$.
    \item  $\forall \boldsymbol{x}: \nabla_{\boldsymbol{x}}|\det\boldsymbol{J}_{f(\boldsymbol{x}, \cdot)}|$ and $\nabla_{\boldsymbol{x}}|\det\boldsymbol{J}_{f^{-1}(\boldsymbol{x}, \cdot)}|$ exist.
    \item (Variability) $\forall u:$ Instances $\boldsymbol{z}_1, \ldots, \boldsymbol{z}_{d+1}$ exist such that $|\det \boldsymbol{M}(u, \boldsymbol{z}_1, \ldots, \boldsymbol{z}_{d+1})| > 0,$ where
    \begin{equation*}
    \small{\boldsymbol{M}(u, \boldsymbol{z}_1, \ldots, \boldsymbol{z}_{d+1})} \triangleq    
    \begin{bsmallmatrix}
    P(u|\boldsymbol{z}_1) & \nabla_u P(u|\boldsymbol{z}_1)\\
    \vdots                & \vdots                 \\
    P(u|\boldsymbol{z}_{d+1}) & \nabla_u P(u|\boldsymbol{z}_{d+1})\\
    \end{bsmallmatrix}
    \end{equation*}
\end{enumerate}
\end{theorem}
}
In the above theorem, $\boldsymbol{Z}$ can be both discrete or continuous. The variability condition implies that $\mathbf{Z}$ must have a strong enough influence on $U$.

\section{Efficient Learning of the BGM}
\label{sec:learning}


\edited{

Given an observed data distribution ($\mathcal{D}$), our goal in this section is to find a BGM $\hat{f}$ that is counterfactually equivalent to the BGM \edited{$f$} underlying the data. First, we formalize the notion of equivalence:

\begin{definition}
\label{def:g-equivalence}
\textit{(Equivalence)}
BGMs $f_1$ and $f_2$ are equivalent iff there exists an invertible function $g(\cdot)$ such that
\begin{align}
\label{eqn:g-equivalence}
    \forall \boldsymbol{x},v:& f_1^{-1}(\boldsymbol{x}, v) = g\Big(f_2^{-1}(\boldsymbol{x}, v)\Big)\text{ or alternatively}\\
\label{eqn:reverse-g-equivalence}
    \forall \boldsymbol{x},u:& f_1(\boldsymbol{x}, u) = f_2\Big(\boldsymbol{x}, g^{-1}(u)\Big).
\end{align}
\end{definition}

\begin{proposition}
\label{prop:counterfactual-equivalence}
BGMs $f_1$ and $f_2$ produce the same counterfactuals iff they are equivalent.
\end{proposition}

If we find an $\hat{f}$ which is equivalent to the BGM \edited{$f$} underlying the data, we can perform Abduction-Action-Prediction (\S\ref{sec:prelim}) using $\hat{f}$ to estimate any counterfactual quantity, which is guaranteed to produce the same counterfactuals as the true BGM \edited{$f$} via Proposition~\ref{prop:counterfactual-equivalence}.

The theorems in \S\ref{sec:identifiability} (see also the lemmas in \Cref{app:sub:lemmas}) provide a tractable objective for learning a BGM that is equivalent to the ground-truth BGM for counterfactual estimation. 
We now describe a recipe for efficiently solving this learning problem.

}

To represent the search space for $\hat{f}$, we use parameterized bijective transforms $\hat{f}_{\theta}(\boldsymbol{x}, \cdot)$ from $\hat{U}$ to $V$, conditioned on $\boldsymbol{X}$.
This resembles transforms used in Conditional Generative Models (CGM).
Furthermore, we require certain constraints on $\hat{f}_{\theta}$ depending on the case, e.g., being strictly monotonic or differentiable, which we take into account to select an appropriate model family and parameterization. For example, both the IV and BC cases require differentiablity of the transform. Conditional Normalizing Flow (CNF)~\cite{papamakarios2021normalizing} models are a good candidate for learning such functions as they are typically built using differentiable mappings with differentiable inverses (diffeomorphisms). Although our approach is applicable to any CGM with the desired properties, we use CNFs in our experiments. Appendix~\ref{app:NF} discusses CNFs in more detail. 


Each set of constraints has an important distributional requirement (the first condition) in the form of (conditional) independence among $\hat{U}$ (\Cref{eqn:reverse_generation_mechanism}) and other variables. One way to guide the learning to attain such conditional independence properties is to use adversarial learning for distribution matching~\cite{li2017alice}.
However, a more elegant solution emerges from flipping the problem.
Instead of passing $(\mathbf{X},V) \in \mathcal{D}$ samples through $\hat{f}_{\theta}^{-1}$ (\Cref{eqn:reverse_generation_mechanism}) and optimizing the transform to satisfy the (conditional) independence, we can sample $\hat{U}$ in a way that satisfies the (conditional) independence needed in each case, and optimize $\hat{f}_{\theta}$ (\cref{eqn:generation_mechanism}) to produce  the observed conditional distribution $P_{\mathcal{D}}(V|\boldsymbol{X})$.
This has two benefits: i) (Conditional) independence is guaranteed by construction. ii) Training objective simply becomes density modeling, which has been greatly studied in the literature.

\textbf{The Markovian Case} in \S\ref{sub:Markovian} provides a simple example to  explain the learning method.
We use a strictly increasing conditional transformation as the search space for $\hat{f}_{\theta}$.
We sample $\hat{U}$ and $\boldsymbol{X}$ independently from a Gaussian distribution and the dataset $\mathcal{D}$, respectively.
This guarantees the first condition of \Cref{thm:intervention}.
We optimize the parameters of $\hat{f}_{\theta}$ to match the conditional distribution $P_{\mathcal{D}}(V|\boldsymbol{X})$.
In doing so, we are free to choose from a diverse set of objectives provided by years of research in generative modeling~\cite{mohamed2016learning}, for instance, variational loss~\cite{kingma2013auto}, adversarial loss~\cite{goodfellow2014generative}, likelihood maximization~\cite{papamakarios2021normalizing}, score matching~\cite{song2019generative} or denoising diffusion~\cite{ho2020denoising}, etc.


\begin{figure}
    \centering
        \begin{subfigure}{0.49\linewidth}
        \centering
        \includegraphics[width=0.6\linewidth]{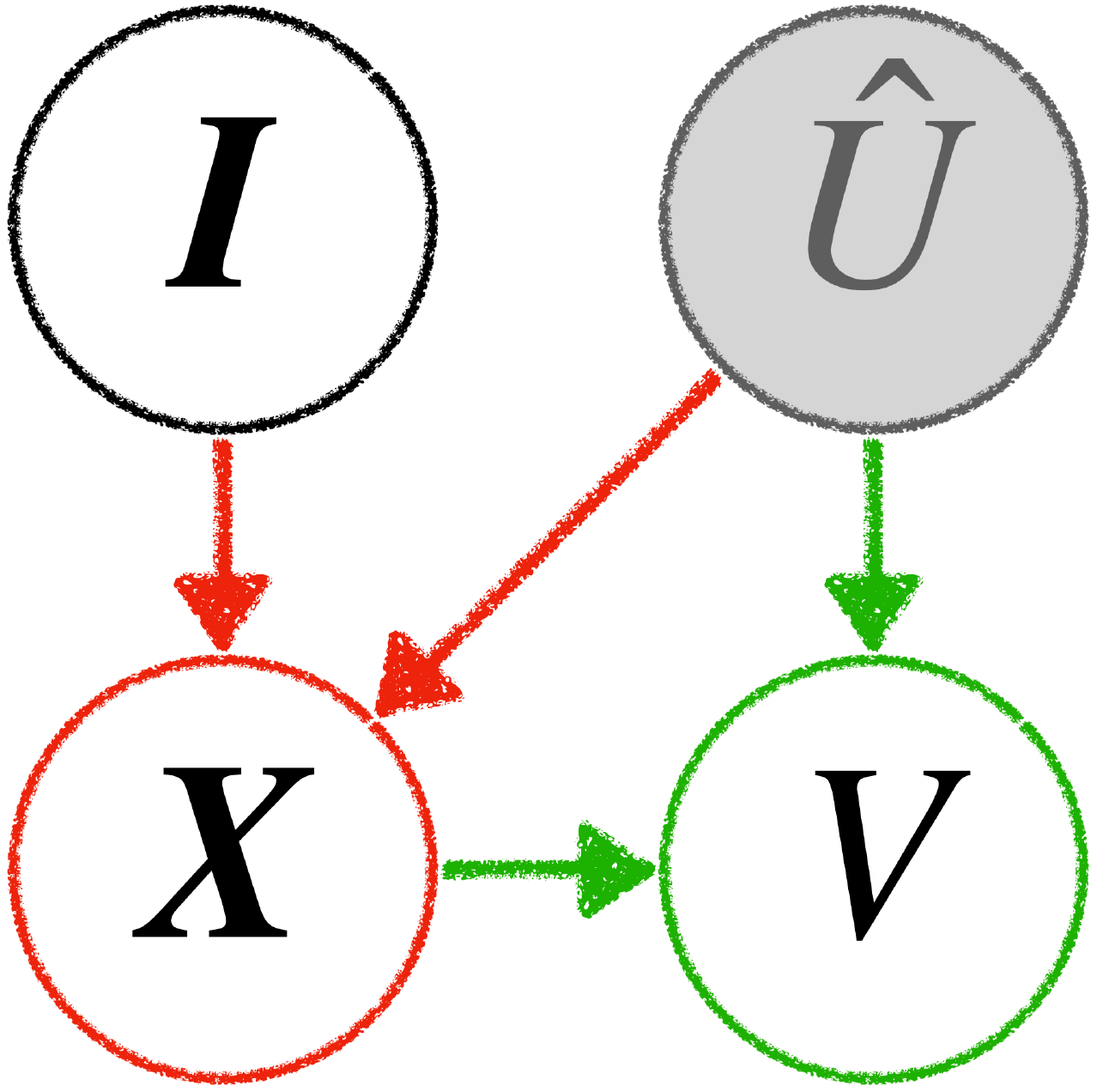}
        \caption{Instrumental Variable}
        \label{subfig:PGM:CNF-IV}
    \end{subfigure}
    \begin{subfigure}{0.49\linewidth}
        \centering
        \includegraphics[width=0.6\linewidth]{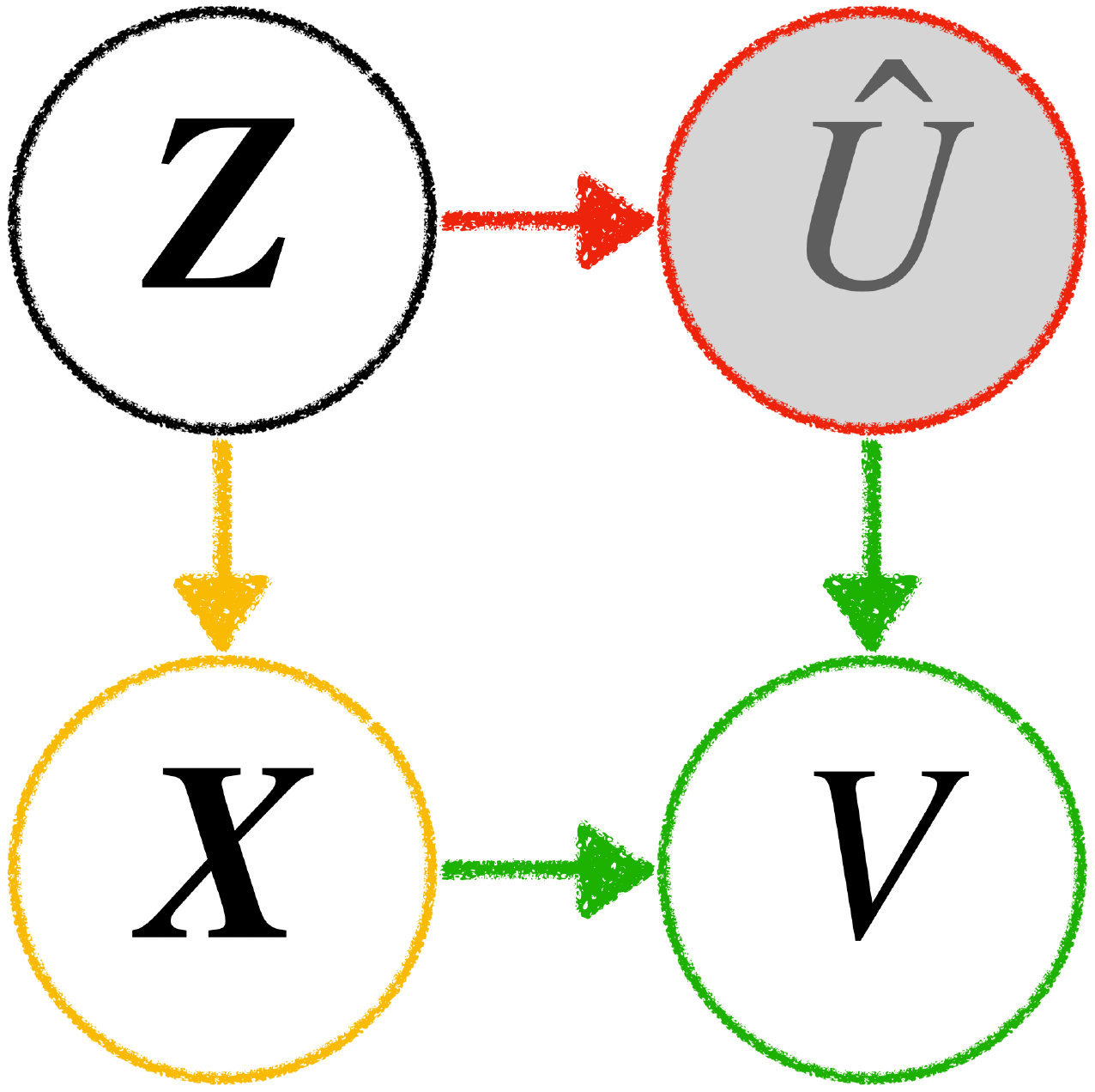}
        \caption{Backdoor Criterion}
        \label{subfig:PGM:CNF-BC}
    \end{subfigure}
    \label{fig:CNF-DGM}

    \begin{subfigure}{0.49\linewidth}
        \centering
        \includegraphics[width=0.95\linewidth]{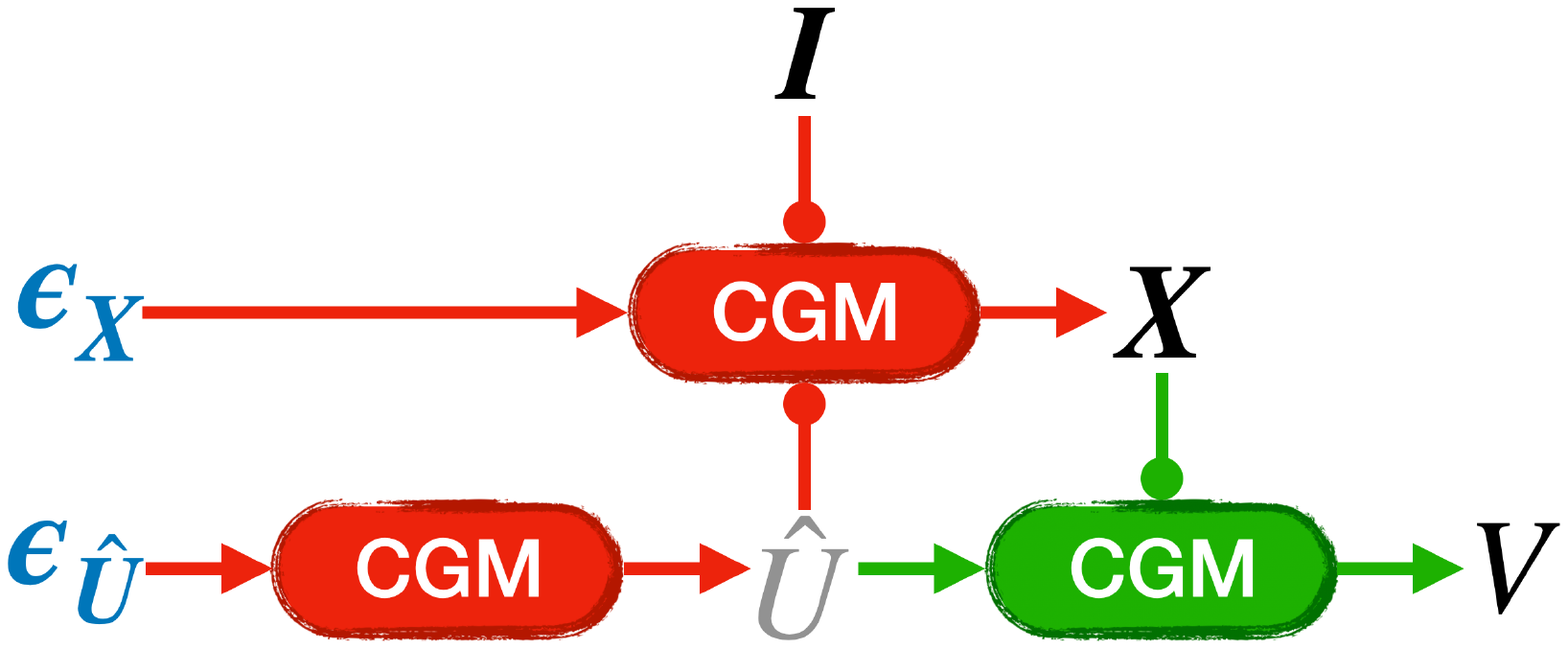}
        \caption{Instrumental Variable}
        \label{subfig:CNF-IV}
    \end{subfigure}
    \begin{subfigure}{0.49\linewidth}
        \centering
        \includegraphics[width=0.95\linewidth]{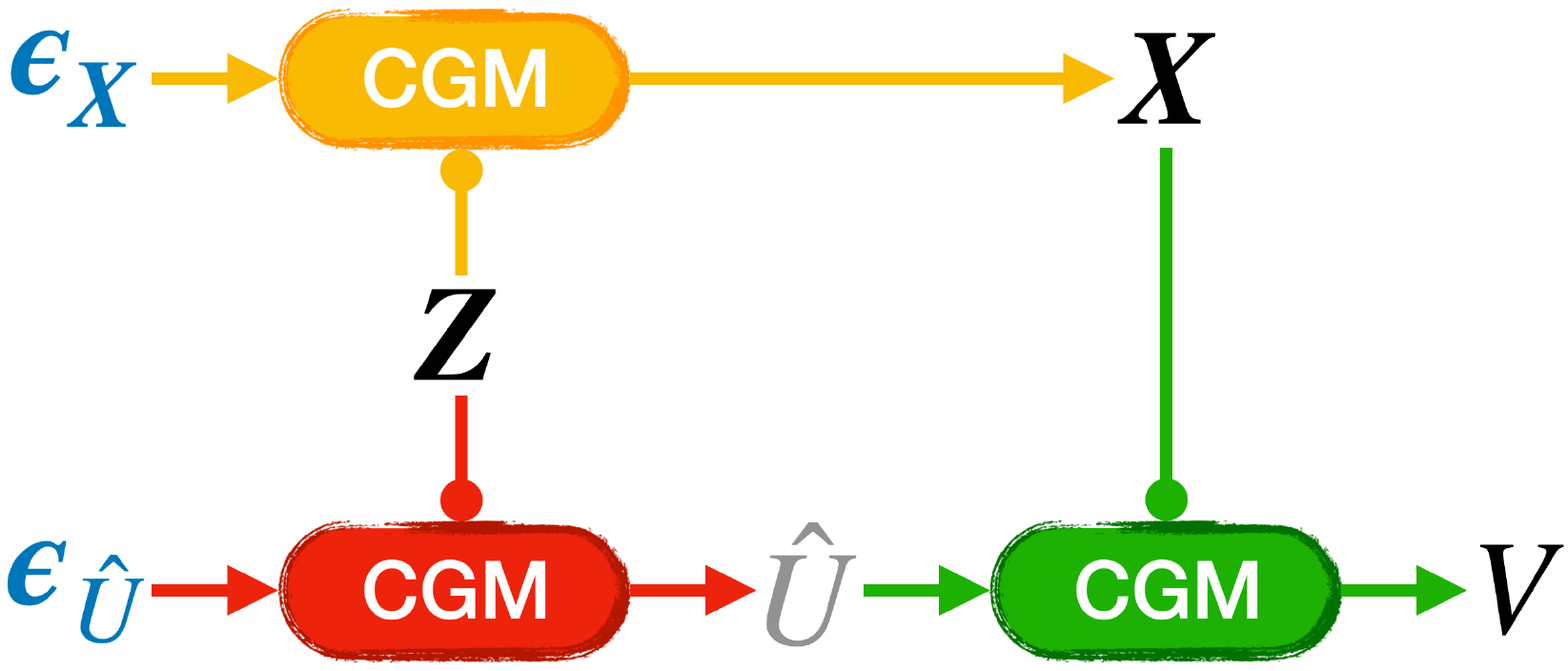}
        \caption{Backdoor Criterion}
        \label{subfig:CNF-BC}
    \end{subfigure}
    
    \caption{Graphical models (top) encode desired (conditional) independencies.
    Structured generative networks (bottom) for learning BGMs are constructed according to the corresponding graphical models.
    Blue variables are sampled independently from Gaussian distributions while black variables are observed in the dataset.
    }
    \label{fig:CNF}
\end{figure}

In the \textbf{IV} (\S\ref{sub:IV}) and \textbf{BC} (\S\ref{sub:BC}) cases on the other hand, training is not as simple at the first glance.
We cannot sample $\hat{U}$ and $\boldsymbol{X}$ independently anymore, as they do not need to be independent.
Additionally, we need to satisfy the conditional independence constraints of each case.

To efficiently encode the conditional independence constraints, we use directed graphical models (DGM)~\cite{koller2009probabilistic} for building \emph{structured generative networks} which consist of CGMs as their building blocks, and inherit all conditional independencies encoded in the DGM.\footnote{Here, we treat DGMs as a pure statistical objects devoid of any causal semantics.}
\Cref{fig:CNF} demonstrates the DGMs used for satisfying the conditional independencies of IV and BC cases,
as well as
structured generative models built according to these DGMs. Note that the shown DGMs and structured generative networks are not the only ones that satisfy the required conditional independence properties. In the BC case, there are several alternatives DGM that encode the same conditional independencies and each define a distinct structured generative network that could be used to learn $\hat{f}$ (see \Cref{app:CI}). Also note that the CGM used for representing $\hat{f}_{\theta}$ (green one) has requirements like monotonicity and differentiability, but the others do not have these restrictions.

\textbf{Training:}
We sample the root variables, $\boldsymbol{\epsilon}_{\boldsymbol{X}}, \epsilon_{\hat{U}}, \boldsymbol{I}$ in \Cref{subfig:CNF-IV}, and $\boldsymbol{\epsilon}_{\boldsymbol{X}}, \epsilon_{\hat{U}}, \boldsymbol{Z}$ in \Cref{subfig:CNF-BC}, independently. 
The epsilon variables (in blue) are sampled from independent Gaussian distributions, and the others from the dataset $\mathcal{D}$.
We train the whole structure end-to-end, with the goal of matching the distribution of $\mathcal{D}$.
This process is akin to a constrained search for $\hat{f}$ where the constraints are embedded in the search space by construction.
The objective function is determined by the choice of CGM.
We can even mix different types of CGMs, and combine different objective functions for their end-to-end optimization.

The end goal of training is to learn the green CGM.
As a result, any CGM whose training does not affect the green CGM can be removed from the network as a simplification.
The yellow CGM in \Cref{subfig:CNF-BC} is such an example, as it is surrounded by black variables that are sampled directly from $\mathcal{D}$, which prevent its gradients from passing through and influencing training of the other CGMs. 
\section{Efficient Counterfactual Estimation}
\label{sec:counterfactual}


Once we have learned a BGM using the techniques in \S\ref{sec:learning}, estimating counterfactual queries is a straightforward application of the Abduction-Action-Prediction procedure (\S\ref{sec:prelim}). We illustrate this using two examples. 

Suppose we are interested in a counterfactual query $V_{\boldsymbol{x}'}|\{\boldsymbol{X} = \boldsymbol{x}, V = v\}$, which seeks to determine the necessity of causation, e.g., given that a patient recovered ($V=1$) under a treatment $X=1$, would she have also recovered without the treatment ($X=0$).  This query is in general non-identifiable from observational ($\mathcal{L}_1$) and interventional ($\mathcal{L}_2$) data~\cite{pearl1999probabilities,tian2000probabilities,li2022probabilities}.
However, if we can approximate the generation mechanism of $V$ as a BGM, and if one of the identifiability conditions of \S\ref{sec:identifiability} holds, there is hope. Specifically, we first learn the BGM from
observational data (\S\ref{sec:learning}). Then, we do the abduction step by inverting the BGM, i.e., $\hat{u} = \hat{f}^{-1}(\boldsymbol{x}, v)$, and  pass the counterfactual $\boldsymbol{x}'$ through the BGM to get the counterfactual estimate, i.e., $v' = \hat{f}(\boldsymbol{x'}, \hat{u})$.

As another example, consider $P(V_{\boldsymbol{x}_2}|\boldsymbol{X} = \boldsymbol{x}_1)$ which questions the effect of treatment on the treated~\cite{shpitser2009effects}.
For the abduction step, we invert the learned BGM ($\hat{f}$) for all the observed samples (e.g., patients) assigned to $\boldsymbol{x}_1$ to obtain samples of the exogenous posterior distribution $P_{\hat{U}|\boldsymbol{x}_1}$ as $\hat{u} = \hat{f}^{-1}(\boldsymbol{x}_1, v), v \sim P(V|\boldsymbol{x_1})$.
For the prediction step, we pass the samples $\hat{u}$ through the BGM with $X=\boldsymbol{x}_2$, to obtain samples of the counterfactual distribution of interest $v' = \hat{f}(\boldsymbol{x}_2, \hat{u}), \hat{u} \sim P_{\hat{U}|\boldsymbol{x}_1}$.

The examples above considered only the generation function of particular node $V \in \boldsymbol{V}$ in the causal diagram.
If all generation functions of an SCM ($\mathcal{F}$) are well-approximated by BGMs, and can also be learned in the settings described in \S\ref{sec:identifiability}, then we can answer every counterfactual query in a similar way.
However, not all generation mechanisms need to be BGMs, or known, for answering a particular counterfactual query of interest.
For instance, there is no need to learn BGMs for ancestors of the variables that we intervene on, or the variables that do not appear in the evidence and do not have dashed-bidirected edges to the evidence.

\section{Experiments}
\label{sec:eval}

We evaluate our techniques in two settings: (i) a simple task that allows us to visualize our approach and prior baselines, and (ii) a real-world video streaming simulation task. 
\Cref{app:exp} has implementation details.

\subsection{Counterfactual Ellipse Generation}
\label{eval:ellipse}

A standard ellipse is determined by two parameters: a semi-major and semi-minor axis ($a, b$).
If we further specify an angle, we get a single point on the ellipse.
Let $U \in \mathbb{R}^2$ be the two parameters of an ellipse, $X \in (0, 2\pi)$ an angle that specifies a single point, $V \in \mathbb{R}^2$ the Cartesian coordinates of this point, and $f$ the function that calculates these coordinates given $U$ and $X$.

Suppose we are given data generated as follows. We sample $z \sim P_Z$, $u \sim P_{U|Z=z}$ and $x \sim P_{X|Z=z}$, and output the data tuple $(z, x, v\coloneqq f(u,x))$. Here, $P_Z$, $P_{U|Z}$, and $P_{X|Z}$ are three predefined distributions (see \Cref{app:ellipse} for details). The important point is that conditioned on $Z$, the ellipse parameters and angles are independent $(U \indep X|Z)$, but $U$ and $X$ are not independent unconditionally. For a specific pair $(x, v)$ observed in the dataset (evidence), our goal is to draw the entire ellipse that the point $v$ belongs to. This can be done by estimating counterfactual queries, $V_{x'}|\{X=x, V=v\}$, for $x'\in (0, 2\pi)$. 





\begin{figure}
\begin{center}
\centerline{\includegraphics[width=0.9\linewidth]{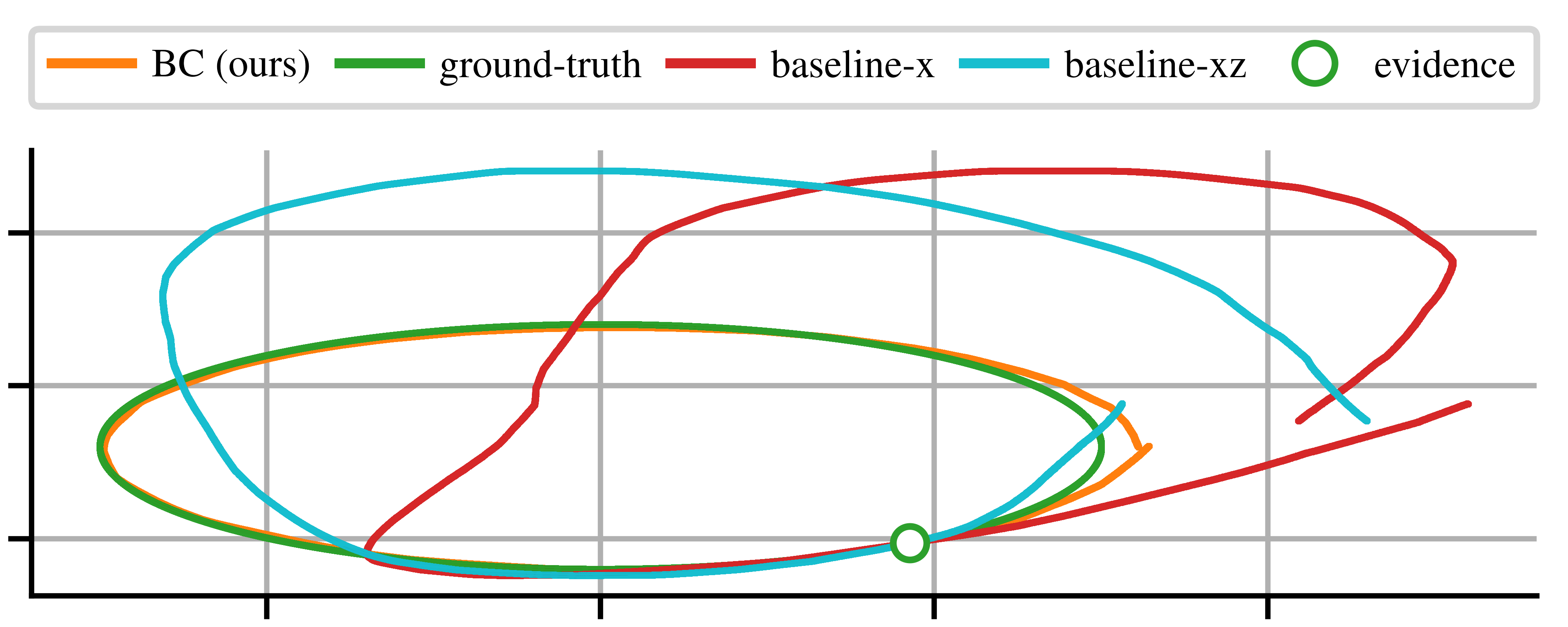}}
\caption{BC (ours) is the only scheme that generates an ellipse.}
\label{fig:ellipse}
\vspace{-7pt}
\end{center}
\end{figure}

This task corresponds to our \textbf{BC} setting with $Z$ as the backdoor variable. We evaluate our method for BC visually (\Cref{fig:ellipse}) and quantitatively (\Cref{table:ellipse}).
As you can see, it achieves a high accuracy as opposed to the baselines, both of which are single CGMs that do not take the causal structure into account, which is common in prior work~\cite{lample2017fader,zhu2017unpaired,he2019attgan}. 
\textbf{Baseline-x} models $P(V|X)$ and \textbf{baseline-xz} models $P(V|X, Z)$.

\begin{table}[t]
\caption{Counterfactual Ellipse Generation Accuracy}
\label{table:ellipse}
\begin{center}
\begin{small}
\begin{sc}
\begin{tabular}{l|ccr}
\toprule
Scheme & BC & Baseline-x & Baseline-xz\\
MAPE & 1\% & 6607\% & 6582\% \\
\bottomrule
\end{tabular}
\end{sc}
\end{small}
\end{center}
\end{table}

\subsection{Case Study: Video Streaming Simulation}
\label{subsec:trace-driven}

Video streaming clients use Adaptive Bitrate (ABR) algorithms to continually adapt the bitrate of a video stream based on network conditions. ABR algorithms select a bitrate for each few-second chunk of video among a finite set of available choices. These algorithms have a significant impact on user experience (e.g., video quality and stalls) and have been the subject of extensive research~\cite{tian2012towards,huang2014buffer,yin2015control,sun2016cs2p,mao2017neural,akhtar2018oboe,spiteri2020bola}. 

Trace-driven simulation is a common approach to design and evaluate ABR algorithms. Here, one collects traces from real video streaming sessions, with each trace providing a timeseries of observed network throughput (and possibly other player metrics) for every video chunk. At simulation time, the traces are replayed (to represent network behavior) while simulating the dynamics of video clients under new ABR algorithms. However, simply replaying a throughput trace can bias simulation outcomes~\cite{bartulovic2017biases, causalsim}, because changing the ABR algorithm could effect the throughput that would have been achieved for the same video streaming session.

\Citet{causalsim} formulated bias removal in this type of simulation as a counterfactual estimation problem where for download of each chunk, $V$ is the achieved throughput, $X$ is the chosen bitrate,  $U$ is the unobserved bottleneck link capacity, and $V = f(U, X)$.
Note that in this problem, $f$ is strictly increasing for each value of $X$ as higher bottleneck link capacity ($U$) increases the achieved throughput ($V$).
They prove identifiability for this counterfactual query assuming data collected in a Randomized Control Trial (RCT) with sufficiently diverse ABR algorithms and low-rank structure of the underlying BGM. 
They use a deterministic auto-encoder~\cite{ghosh2019variational} equipped with adversarial learning for distribution matching for counterfactual estimation, and present experimental results (including a real-world ABR design case study) that show it significantly improves simulation accuracy compared to standard (biased) trace-driven simulators. 

Next, we demonstrate how each set of conditions for identifiability (\S\ref{sec:identifiability}) translates to this real-world problem along with the efficacy of our practical algorithm (\S\ref{sec:learning}) using the ABR simulator provided by \citet{causalsim}. We use the simulator to generate traces for each setting, which we use to learn the BGM.
The simulator gives us ground-truth counterfactuals for error calculation.
We normalize the Mean Squared Error (MSE) of our method's counterfactual estimates by the MSE of a standard (biased) video streaming simulator that assumes chosen bitrates do not affect the achieved throughput.
All results are in \Cref{table:abr} with mean and standard deviation calculated over ten random seeds trained until training loss convergence.

\begin{table}
\caption{Counterfactual estimation accuracy of different schemes and their training time for video streaming simulation}
\label{table:abr}
\begin{center}
\begin{small}
\begin{sc}
\begin{tabular}{lcr}
\toprule
Scheme & Normalized MSE $(\%)$ & Time ($s$) \\
\midrule
Markovian    & \; 4.1 $\pm$ 2.1& \; \; 28 $\pm$ 0\\
IV    & 10.0 $\pm$ 0.7& \; 127 $\pm$ 1\\
CausalSim    & \; 9.3 $\pm$ 1.0& 1091 $\pm$ 1\\
BC    & 10.4 $\pm$ 7.8& \; \; 37 $\pm$ 8\\
IV+BC    & \; 6.4 $\pm$ 2.4& \; \; 49 $\pm$ 1\\
\bottomrule
\end{tabular}
\end{sc}
\end{small}
\end{center}
\end{table}

\textbf{The Markovian Case}:
The causal structure in this problem is not Markovian because $X$ (chosen bitrate) and $V$ (observed throughput) are both affected by underlying network conditions (latent confounder). 
However, we can remove the confounding effect if the ABR algorithm used for trace collection chooses random bitrates from time to time, and we only select the subset of traces with these random decisions as $\mathcal{D}$ (\Cref{fig:causal-dag}).
This scheme achieves the lowest error in Table~\ref{table:abr} as it has access to the most pristine data (perfect confounding removal by invasive randomization).

\textbf{Instrumental Variable (IV)}:
It is common for video service providers to conduct RCTs over various ABR algorithms for their comparison.
If traces are collected from an RCT, their causal structure naturally fits \Cref{subfig:IV} where IV ($I$) is the algorithm identifier.
This is the same setting that \textbf{CausalSim}~\yrcite{causalsim} explores, so we use their adversarial learning method as a baseline.
We collect RCT data over ten different provided algorithms.
Our IV method achieves almost the same accuracy as CausalSim (slight difference is not statistically significant), but converges $8.6 \times$ faster since it does not require bi-level (adversarial) optimization.

\textbf{Backdoor Criterion (BC)}:
Buffer based ABR algorithms are those that only make use of the client's current playback buffer level to make bitrate decisions.
They are strong ABR algorithms despite their simplicity and are widely used in practice~\cite{yan2020learning}.
If a trace is collected using a buffer based algorithm (and includes playback buffer observations), the causal structure follows \Cref{subfig:BC} where $Z$ is the buffer used for choosing the bitrate. Hence, it satisfies the backdoor criterion. Table~\ref{table:abr} shows that our BC method applied to buffer-based traces is quite accurate even without access to RCT data (although it is slightly worse than CausalSim and IV). 


\textbf{IV + BC}:
It is possible to combine settings explored in \S\ref{sec:identifiability} to further improve accuracy provided the problem has the appropriate causal structure. For example, in video streaming simulation, if $\mathcal{D}$ is collected in an RCT over buffer based algorithms, the BC and IV cases apply simultaneously. We refer to this case as IV+BC. 
Similar to the IV case, the algorithm identifier is the instrumental variable while the tuple (algorithm identifier, buffer) satisfies the backdoor criterion. \Cref{app:video:IV+BC} describes the structured generative network for this case. We apply it to traces collected using an RCT over two buffer based algorithms. 
IV+BC decreases the average error compared to the BC scheme by almost $38\%$.
\section{Concluding Remarks}
\label{sec:conclusion}

In this work, we defined Bijective Generation Mechanisms (BGM), a class of models that contain several widely used causal models in the literature (\S\ref{sec:problem}).
We established their counterfactual identifiability for three well-known causal structures (\S\ref{sec:identifiability}) and proposed a practical learning method that casts learning a BGM as structured generative modeling (\S\ref{sec:learning}).
We evaluated our methodology in a visual task and demonstrated its application to a real-world video streaming simulation task (\S\ref{sec:eval}).
Finite sample analysis of our identifiability theorems, extending the identifiable causal structures, and applications of the proposed method to real-world problems in various field, e.g., econometrics, computer systems, causal ML, etc., are exciting directions for future work.
\section*{Acknowledgements}
\label{sec:ack}

We would like to thank Behrooz Tahmasebi for discussions that led to the proof of \Cref{thm:IV}.
This work was supported by NSF grant 1751009, a gift from the CSAIL SystemsThatLearn (STL) Initiative, and Google, Intel, and Amazon as part of the MIT Data Systems and AI (DSAIL) lab.

\bibliography{paper}
\bibliographystyle{icml2023}

\newpage
\appendix
\onecolumn

\section{Neural Methods for Causal Estimation}
\label{app:related}
\citet{xia2021causal} uses Neural Causal Models (NCMs) learned from observational data for identification and estimation of interventional $(\mathcal{L}_2)$ queries, assuming knowledge of the underlying causal diagram (potentially non-Markovian) and discrete endogenous variables.

\citet{kocaoglu2018causalgan} and \citet{zevcevic2021relating} use adversarial training and Graph Neural Networks (GNNs)~\cite{wu2020comprehensive}, respectively, to learn implicit SCMs from observational data, assuming knowledge of the causal diagram and the Markovian assumption.
Learned SCMs are then used for answering interventional queries $(\mathcal{L}_2)$, which are known to be identifiable given the Markovianity assumption~\citep[Corol.~2]{bareinboim2022pearl}. 

\citet{pawlowski2020deep} and \citet{sanchez2021vaca} use deep conditional generative models of various forms, structured according to the known causal diagram, to learn the SCM from observational data assuming no unobserved confounding (Markovain SCM).
The learned SCMs are further used for interventional and counterfactual estimation.
\citet{sanchez2021diffusion} uses diffusion denoising probabilistic models to learn conditional distribution of images given labeled attributes, which are further used for counterfactual image generation.
However, they do not have any identifiability analysis. 
In fact \citet{non-ident} shows counterfactual non-identifiability of generation mechanisms of multi-dimensional variables from observational data in Markovian settings.
To assess the quality of non-identifiable image counterfactuals in Markovian SCMs, \citet{monteiro2023measuring} revisits axiomatic definition of counterfactuals by measuring their \textit{composition}, \textit{reversibility}, and \textit{effectiveness}.

\citet{xia2022neural} use Generative Adversarial Networks (GANs)~\cite{goodfellow2014generative} to learn proxy SCMs from observational and interventional data, assuming knowledge of (non-Markovian) causal diagram, and discrete endogenous variables.
They utizile the proxy SCM for counterfactual $(\mathcal{L}_3)$ identification and estimation.
Gumbel-Max SCMs~\cite{oberst2019counterfactual} and their generalizations~\cite{lorberbom2021learning} have been used for counterfactual estimation of categorical variables from observational data.
However, we focus mostly on continuous domains.

Assuming SCMs with additive noise (ANM)~\cite{hoyer2008nonlinear} and Markovianity, \citet{geffner2022deep} learns both the underlying causal structure $\mathcal{G}$ and SCM from observational data, and uses them for estimation of interventional and counterfactual queries.
However, their identifiability analysis is restricted to interventional queries only.
Non-linear ANMs are a special case of the class of SCMs we consider in this work. Furthermore, we allow existence of unobserved confounders, and prove counterfactual identifiability of our models.
\citet{hartford2017deep} uses two-stage supervised learning methods to estimate counterfactual queries using Instrumental Variables (IVs)~\cite{angrist1996identification}, assuming SCMs with ANMs.
\citet{causalsim} proves identifiability of counterfactual queries from Randomized Control Trial (RCT) data which is a special IV case, assuming low-rank generation mechanisms. 
Furthermore, they utilize a deterministic auto-encoder~\cite{ghosh2019variational} equipped with adversarial learning for distribution matching to enable efficient counterfactual estimation.
Our treatment does not need any assumptions about the rank of generation mechanisms.

\citet{khemakhem2021causal} uses Affine Causal Autoregressive Flows~\cite{dinh2016density} for learning the underlying causal structure $\mathcal{G}$ from observational data, assuming absence of unobserved confounder.
They prove identifiability of interventional queries assuming Location Scale Noise Models (LSNM)~\cite{strobl2022identifying}.
Additionally, they propose using the learned SCM for counterfactual estimation, without any identifiability analysis.
We allow unobserved confounders to exist, do not restrict generation functions to LSNMs, and also prove counterfactual identifiability.

\citet{louizos2017causal} uses Variational Auto-Encoders (VAE)~\cite{kingma2013auto} to estimate counterfactual queries in cases where sufficient proxy variables~\cite{carroll2006measurement,kuroki2014measurement,miao2018identifying,wang2019blessings,lee2021causal} of unobserved confounders are available for identifiability.
\citet{johansson2016learning,shalit2017estimating,yao2018representation} learn representations for estimating \emph{Individual Treatment Effect (ITE)} assuming \emph{strong ignorability} and binary treatments.
Our work is not limited to discrete domains, and is not limited to specific counterfactual quantities like ITE.

\section{Proofs}
\label{app:proof}
\subsection{\Cref{prop:counterfactual-equivalence}}
\label{app:proof:prop:counterfactual-equivalence}
\begin{equation}
    \text{BGMs }f_1, f_2\text{ produce the same counterfactuals }\iff f_1, f_2\text{ are equivalent BGMs}
\end{equation}
\begin{proof}
First, we prove $\Leftarrow$.
Consider an arbitrary counterfactual query $V_{\boldsymbol{x'}}|\boldsymbol{X} = \boldsymbol{x}, V = v$.
In the abduction step, both BGMs use the evidence $\boldsymbol{X} = \boldsymbol{x}, V = v$ to infer the exogenous variable $U$.
We refer to $f_1$'s inferred exogenous variable as $u_1$ and $f_2$'s inferred exogenous variable as $u_2$.
Using \Cref{eqn:g-equivalence} we have 
\begin{equation}
    u_1 = g(u_2).
\end{equation}
In the prediction step, $f_1$ and $f_2$ give $f_1(\boldsymbol{x'}, u_1) = f_1\Big(\boldsymbol{x'}, g(u_2)\Big)$ and $f_2(\boldsymbol{x'}, u_2)$ as their counterfactual estimates, respectively.
Using \Cref{eqn:reverse-g-equivalence} we get
\begin{equation}
    f_1\Big(\boldsymbol{x'}, g(u_2)\Big) = f_2(\boldsymbol{x'}, u_2).
\end{equation}
Hence, estimated counterfactuals are equal.

Next, we prove $\Rightarrow$.
Due to both $f_1$ and $f_2$ being BGMs, we can easily verify the following relationship holds between them:
\begin{equation}
\label{app:eqn:bgm-relation}
    \forall \boldsymbol{x},u_1: f_1(\boldsymbol{x}, u_1) = f_2\Big(\boldsymbol{x}, g^{-1}(\boldsymbol{x}, u_1)\Big)
\end{equation}
where
\begin{equation}
    \forall \boldsymbol{x},u_1: g^{-1}(\boldsymbol{x}, u_1) = f_2^{-1}\bigg(\boldsymbol{x}, \Big(f_1(\boldsymbol{x}, u_1)\Big)\bigg).
\end{equation}
Now suppose we use both $f_1$ and $f_2$ for estimating the counterfactual query $V_{\boldsymbol{x'}}|\boldsymbol{X} = \boldsymbol{x}, V = v$.
$f_1$'s estimate would be $f_1(\boldsymbol{x'}, u_1)$. 
Using \Cref{app:eqn:bgm-relation}, this estimate is equal to $f_2\Big(\boldsymbol{x'}, g^{-1}(\boldsymbol{x'}, u_1)\Big)$.
Using $f_2$ for estimating the same query, it infers the exogenous variable $g^{-1}(\boldsymbol{x}, u_1)$ in the abduction step.
In the prediction step, its counterfactual estimate would be $f_2\Big(\boldsymbol{x'}, g^{-1}(\boldsymbol{x}, u_1)\Big)$.
As both the counterfactual estimates are equal, we have
\begin{equation}
    \forall \boldsymbol{x'}, \boldsymbol{x}, u_1: f_2\Big(\boldsymbol{x'}, g^{-1}(\boldsymbol{x'}, u_1)\Big) = f_2\Big(\boldsymbol{x'}, g^{-1}(\boldsymbol{x}, u_1)\Big).
\end{equation}
Using invertibility property of the BGM $f_2$ we get
\begin{equation}
    \forall \boldsymbol{x'}, \boldsymbol{x}, u_1: g^{-1}(\boldsymbol{x'}, u_1) = g^{-1}(\boldsymbol{x}, u_1) \rightarrow
    \forall \boldsymbol{x'}, \boldsymbol{x}, u_1: g\Big(\boldsymbol{x'}, g^{-1}(\boldsymbol{x}, u_1)\Big) = u_1.
\end{equation}
The only way the last equality could hold for all possible $\boldsymbol{x}, \boldsymbol{x'}$ is if $g^{-1}$ does not depend on its first argument, i.e.,
\begin{equation}
    g^{-1}(\boldsymbol{x}, u_1) = g^{-1}(u_1),
\end{equation}
which means that $f_1$ and $f_2$ are equivalent.
\end{proof}
\begin{remark}
The reason why we have this indeterminacy $g(\cdot)$ is partly due to the fact that the prior distribution over exogenous variables $\Big(P(U)\Big)$ is unknown.
Each choice of this prior distribution would result in a different $g(\cdot)$.
\end{remark}

\subsection{Counterfactual Equivalence Lemmas}
\label{app:sub:lemmas}
In this section, we present three lemmas that are essential for the proof of counterfactual identifiability results in \S\ref{sec:identifiability}.
\subsubsection{The Markovian Case}
\begin{lemma}
\label{lemma:intervention}
BGMs $f$ and $\hat{f}$ that produce the same distribution $P_{\mathcal{D}}(\boldsymbol{X}, V)$ are equivalent if
\begin{enumerate}
    \item (Markovian) $U \indep \boldsymbol{X}$ and $\hat{U} \indep \boldsymbol{X}$.
    \item for all $\boldsymbol{x}$, $f(\boldsymbol{x}, \cdot)$ and $\hat{f}(\boldsymbol{x}, \cdot)$ are either strictly increasing or strictly decreasing functions.
\end{enumerate}
\end{lemma}
\begin{proof}
We will show that BGMs $f$ and $\hat{f}$ produce the same counterfactuals.
Using \Cref{prop:counterfactual-equivalence}, we conclude their equivalence.
Suppose we are interested in the counterfactual query $V_{\boldsymbol{x'}}|\boldsymbol{X} = \boldsymbol{x}, V = v$.
Without loss of generality, consider only $\boldsymbol{x}, v$ samples for which $P_{\boldsymbol{X}, V} (\boldsymbol{x}, v) > 0$.
Let $F(\boldsymbol{x}, v) \coloneqq P(V \le v | \boldsymbol{X} = \boldsymbol{x})$ be the conditional Cumulative Distribution Function (CDF) and $F^{-1}(\boldsymbol{x}, \alpha)$ the quantile function, which exists where $P_{\boldsymbol{X}, V} (\boldsymbol{x}, v) > 0$.   
In the abduction step of counterfactual estimation, both BGMs $f$ and $\hat{f}$ will return the $F(\boldsymbol{x}, v)^{\text{th}}$ or $\Big(1 - F(\boldsymbol{x}, v)^{\text{th}}\Big)$ quantile of their corresponding exogenous distribution $\Big(P(U|\boldsymbol{X}) \text{ and }P(\hat{U}|\boldsymbol{X})\Big)$ if they are both increasing, or decreasing, respectively.
These quantiles might in fact be two distinct values.
As $U, \hat{U} \indep \boldsymbol{X}$, the quantile function of $U, \hat{U}$ given $\boldsymbol{X}$ is independent of $\boldsymbol{X}$, and equal to the quantile function of the marginal $U, \hat{U}$.
Hence, the action step would not change the estimated quantile.
In the prediction step, both BGMs $f$ and $\hat{f}$ would estimate the same value $F^{-1}\Big(\boldsymbol{x'}, F(\boldsymbol{x}, v)\Big)$ if increasing, or other similar quantile variations if decreasing.
\end{proof}

\subsubsection{Instrumental Variable (IV)}
\begin{lemma}
\label{lemma:IV}

For $\boldsymbol{X} \in \mathbb{X} \triangleq \{\boldsymbol{x}_1, \ldots, \boldsymbol{x}_n\}$ and $\boldsymbol{I} \in \mathbb{I} \triangleq \{\boldsymbol{i}_1, \ldots, \boldsymbol{i}_n\}$, BGMs $f$ and $\hat{f}$ that produce the same distribution $P_{\mathcal{D}}(\boldsymbol{X}, V)$ are equivalent if
\begin{enumerate}
    \item (IV) $\boldsymbol{I} \indep U$ and $\boldsymbol{I} \indep \hat{U}$.
    \item  for all $\boldsymbol{x} \in \mathbb{X}$, $f^{-1}(\boldsymbol{x}, \cdot)$ and $\hat{f}(\boldsymbol{x}, \cdot)$ are either strictly increasing or strictly decreasing, and two times differentiable.
    \item $P_{\hat{U}}(\cdot)$ is differentiable.
    \item $P_{\mathcal{D}}(\boldsymbol{i}, \boldsymbol{x}, \cdot)$ is differentiable for every $\boldsymbol{i}, \boldsymbol{x}$.
    \item (Positivity) $\forall u, \hat{u}, \boldsymbol{x} \in \mathbb{X}: P_{U, \boldsymbol{X}}(u, \boldsymbol{x}) > 0$ and $P_{\hat{U}, \boldsymbol{X}}(\hat{u}, \boldsymbol{x}) > 0$.
    \item (Variability) $\forall u: |\det \boldsymbol{M}_{\mathcal{D}}(u, \mathbb{I})| \ge c$, where $c$ is a positive constant and 
    \begin{equation}
    \boldsymbol{M}_{\mathcal{D}}(u, \mathbb{I}) \triangleq    
    \begin{bmatrix}
    P_{\mathcal{D}}(\boldsymbol{x}_1|u,\boldsymbol{i}_1) & \ldots & P_{\mathcal{D}}(\boldsymbol{x}_n|u,\boldsymbol{i}_1) \\
    \vdots                   & \ddots & \vdots                   \\
    P_{\mathcal{D}}(\boldsymbol{x}_1|u,\boldsymbol{i}_n) & \ldots & P_{\mathcal{D}}(\boldsymbol{x}_n|u,\boldsymbol{i}_n) \\
    \end{bmatrix}
    \end{equation}
\end{enumerate}
\end{lemma}

\begin{proof}
Define Cumulative Distribution Functions (CDF)  $k(u) = P(U \le u)$, $\hat{k}(\hat{u}) = P(\hat{U} \le \hat{u})$.
We use the CDFs to transform random variables into uniform distributions between 0 and 1.
Define $Z = k(U)$, $\hat{Z} = \hat{k}(\hat{U})$. 
Due to Probability Integral transform we have $Z, \hat{Z} \sim Unif(0, 1)$. 
Furthermore $\forall \boldsymbol{x} \in \mathbb{X}: Z = s(\boldsymbol{x},\hat{Z})$ where $s(\boldsymbol{x},\cdot) = k(\cdot) \circ f^{-1}(\boldsymbol{x},\cdot) \circ \hat{f}(\boldsymbol{x},\cdot) \circ \hat{k}^{-1}(\cdot)$.
Because $U \indep \boldsymbol{I}$ (The first condition), and because $Z$ is a deterministic function of $U$, we conclude $Z \indep \boldsymbol{I}$. 
Using a similar argument, $\hat{Z} \indep \boldsymbol{I}$.
\begin{align}
    \forall \hat{z}, \boldsymbol{i}&: P_{\hat{Z}|\boldsymbol{I}}(\hat{z}|\boldsymbol{i}) = 1\\
    \rightarrow \forall \hat{z}, \boldsymbol{i}&: \sum_{\ell=1}^n P_{\hat{Z},\boldsymbol{X}|\boldsymbol{I}}(\hat{z},\boldsymbol{x}_{\ell}|\boldsymbol{i})=1\\
    \rightarrow \forall \hat{z}, \boldsymbol{i}&: \sum_{\ell=1}^n P_{\hat{Z}|\boldsymbol{X},\boldsymbol{I}}(\hat{z}|\boldsymbol{x}_{\ell},\boldsymbol{i})P_{\boldsymbol{X}|\boldsymbol{I}}(\boldsymbol{x}_{\ell}|\boldsymbol{i})=1
\end{align}
Using conditions $2, 3, 4, 6$ we know that $\forall \boldsymbol{x} \in \mathbb{X}: s(\boldsymbol{x},\cdot)$ is strictly increasing and two times differentiable.
Hence, using the change of variable formula we have:

\begin{equation}
\label{eqn:2}
    \rightarrow \forall \hat{z}, \boldsymbol{i}: \sum_{\ell=1}^n P_{Z|\boldsymbol{X},\boldsymbol{I}}(s(\boldsymbol{x}_{\ell},\hat{z})|\boldsymbol{x}_{\ell},\boldsymbol{i})P_{\boldsymbol{X}|\boldsymbol{I}}(\boldsymbol{x}_{\ell}|\boldsymbol{i})\frac{\partial s(\boldsymbol{x}_{\ell},\hat{z})}{\partial \hat{z}}=1
\end{equation}
Using $Z \indep {\boldsymbol{I}}$ we have:
\begin{equation}
    P_{Z|\boldsymbol{X},\boldsymbol{I}}(z|\boldsymbol{x},\boldsymbol{i})P_{\boldsymbol{X}|\boldsymbol{I}}(\boldsymbol{x}|\boldsymbol{i}) = P_Z (z)P_{\boldsymbol{X}|Z,\boldsymbol{I}}(\boldsymbol{x}|z,\boldsymbol{i})
\end{equation}
$Z \sim Unif(0,1)$ thus
\begin{equation}
\label{eqn:3}
    \rightarrow P_{Z|\boldsymbol{X},\boldsymbol{I}}(z|\boldsymbol{x},\boldsymbol{i})P_{\boldsymbol{X}|\boldsymbol{I}}(\boldsymbol{x}|\boldsymbol{i}) = P_{\boldsymbol{X}|Z,\boldsymbol{I}}(\boldsymbol{x}|z,\boldsymbol{i})
\end{equation}
Now we combine \Cref{eqn:2} and \Cref{eqn:3}:
\begin{equation}
\label{eqn:4}
    \rightarrow \forall \hat{z}, \boldsymbol{i}: \sum_{\ell=1}^n P_{\boldsymbol{X}|Z,\boldsymbol{I}}(\boldsymbol{x}_{\ell}|s(\boldsymbol{x}_{\ell},\hat{z}), \boldsymbol{i})\frac{\partial s(\boldsymbol{x}_{\ell},\hat{z})}{\partial \hat{z}}=1
\end{equation}
We can write \Cref{eqn:4} in vector product form as:
\begin{equation}
\forall \hat{z}, \boldsymbol{i}: 
\begin{bmatrix}
    P_{\boldsymbol{X}|Z,\boldsymbol{I}}(\boldsymbol{x}_1|s(\boldsymbol{x}_1,\hat{z}), \boldsymbol{i}) & \ldots & P_{\boldsymbol{X}|Z,\boldsymbol{I}}(\boldsymbol{x}_n|s(\boldsymbol{x}_n,\hat{z}), \boldsymbol{i}) & -1
\end{bmatrix}
\begin{bmatrix}
    \frac{\partial s(\boldsymbol{x}_1,\hat{z})}{\partial \hat{z}} \\
    \vdots \\
    \frac{\partial s(\boldsymbol{x}_n,\hat{z})}{\partial \hat{z}} \\
    1
\end{bmatrix}
=0
\end{equation}
Now we combine all $n$ equations of this form for $\boldsymbol{i} \in \mathbb{I}$ in matrix format:
\begin{equation}
\label{eqn:5}
\forall \hat{z}:
\begin{bmatrix}
    p_{\boldsymbol{X}|Z,\boldsymbol{I}}(\boldsymbol{x}_1|s(\boldsymbol{x}_1,\hat{z}), \boldsymbol{i}_1) & \ldots & p_{\boldsymbol{X}|Z,\boldsymbol{I}}(\boldsymbol{x}_n|s(\boldsymbol{x}_n,\hat{z}), \boldsymbol{i}_1) & -1\\
    \vdots                               & \vdots & \vdots                               & \vdots\\
    p_{\boldsymbol{X}|Z,\boldsymbol{I}}(\boldsymbol{x}_1|s(\boldsymbol{x}_1,\hat{z}), \boldsymbol{i}_n) & \ldots & p_{\boldsymbol{X}|Z,\boldsymbol{I}}(\boldsymbol{x}_n|s(\boldsymbol{x}_n,\hat{z}), \boldsymbol{i}_n) & -1
\end{bmatrix}
\begin{bmatrix}
    \frac{\partial s(\boldsymbol{x}_1,\hat{z})}{\partial \hat{z}} \\
    \vdots \\
    \frac{\partial s(\boldsymbol{x}_n,\hat{z})}{\partial \hat{z}} \\
    1
\end{bmatrix}
=
\begin{bmatrix}
    0\\
    \vdots \\
    0
\end{bmatrix}
\end{equation}

As argued above, $\forall \boldsymbol{x} \in \mathbb{X}: s(\boldsymbol{x},\cdot)$ is strictly increasing and two times differentiable.
Combining this with the positivity assumption we have
\begin{equation}
\label{eqn:6}
    \forall \boldsymbol{x} \in \mathbb{X}: s(\boldsymbol{x},0) = 0
\end{equation}
Consider \Cref{eqn:5} in $\hat{z}=0$.
The first matrix's rank is $n$ because of the variability condition, and has $n+1$ columns. 
So its nullspace has the form 
$k\begin{bmatrix} 1 & \cdots & 1 \end{bmatrix}^\intercal$.
This implies that 
\begin{equation}
\label{eqn:7}
    \forall \boldsymbol{x} \in \mathbb{X}: \frac{\partial s(\boldsymbol{x},\hat{z})}{\hat{z}}|_{\hat{z}=0} = 1.
\end{equation}
Next, we divide $[0,1]$ into $N$ pieces. 
We prove by induction on $m$ that for large enough $N$
\begin{equation}
\label{eqn:8}
    \forall m \in \{1,\ldots,N\}, \boldsymbol{x} \in \mathbb{X}: |s(\boldsymbol{x},\frac{m}{N})-\frac{m}{N}| \le \frac{m}{2N^2}B
\end{equation}
where $B = \max\limits_{\boldsymbol{x},\hat{z}} \frac{\partial^2s(\hat{z},\boldsymbol{x})}{\partial \hat{z}}$.

\noindent For $m=1$, using Taylor's theorem we have 
\begin{equation}
    \exists \xi \in [0,\frac{1}{N}]:
    s(\boldsymbol{x},\frac{1}{N}) = s(\boldsymbol{x},0) + \frac{1}{N}\frac{\partial s(\boldsymbol{x},\hat{z})}{\partial \hat{z}}|_{\hat{z}=0} + \frac{1}{2N^2}\frac{\partial^2 s(\boldsymbol{x},\hat{z})}{\partial \hat{z}}|_{\hat{z}=\xi}
\end{equation}
Using \Cref{eqn:6,eqn:7}
\begin{align}
    \rightarrow \exists \xi \in [0,\frac{1}{N}]:
    s(\frac{1}{N}, \boldsymbol{x}) &= \frac{1}{N} + \frac{1}{2N^2}\frac{\partial^2 s(\boldsymbol{x},\hat{z})}{\partial \hat{z}}|_{\hat{z}=\xi}\\
    \rightarrow |s(\boldsymbol{x},\frac{1}{N}) - \frac{1}{N}| &\le \frac{B}{2N^2}
\end{align}

\noindent Now suppose that \Cref{eqn:8} holds for $m$, we will show that it holds for $m+1$ too.
We know that $s(\boldsymbol{x},\cdot)$ is differentiable.
Combining it with assumption 6 implies that each element of matrix $M$ is differentiable with respect to $\hat{z}$. 
Furthermore, determinant is a polynomial function of all elements of $M$ which is differentiable with respect to every element. 
Thus determinant of $M$ is differentiable with respect to $\hat{z}$. 
This means that if we perturb $\hat{z}$ in each element of $M$ by a sufficiently small amount, $M$ will remain full rank. 
As a result, for a large enough $N$ the left matrix in \Cref{eqn:5}'s rank is still $n$, and its null space is still one-dimensional. 
This means that 
\begin{equation}
\label{eqn:9}
    \forall \boldsymbol{x} \in \mathbb{X}: \frac{\partial s(\boldsymbol{x},\hat{z})}{\hat{z}}|_{\hat{z}=\frac{m}{N}} = 1.
\end{equation}
using Taylor's theorem we have 
\begin{equation}
    \exists \xi \in [\frac{m}{N},\frac{m+1}{N}]:
    s(\boldsymbol{x},\frac{m+1}{N}) = s(\boldsymbol{x},\frac{m}{N}) + \frac{1}{N}\frac{\partial s(\boldsymbol{x},\hat{z})}{\partial \hat{z}}|_{\hat{z}=\frac{m}{N}} + \frac{1}{2N^2}\frac{\partial^2 s(\boldsymbol{x},\hat{z})}{\partial \hat{z}}|_{\hat{z}=\xi}
\end{equation}
Using \Cref{eqn:8} for $m$ and \Cref{eqn:9} we conclude that
\begin{equation}
    |s(\boldsymbol{x},\frac{m+1}{N})-\frac{m+1}{N}| \le \frac{m+1}{2N^2}B
\end{equation}
This concludes the proof of \Cref{eqn:8} by induction for all values of $m \in \{1,\cdots,N\}$.
In other words, function $s(\boldsymbol{x},\cdot)$ can get as close as wanted to identity in all points $\frac{m}{N}$. 
Using this and the fact that $\forall \boldsymbol{x} \in \mathbb{X}: s(\boldsymbol{x},\cdot)$ is differentiable implies that $\forall \boldsymbol{x} \in \mathbb{X}: s(\boldsymbol{x},\hat{z}) = \hat{z}$. 
This concludes the proof with $g(\cdot) =   k(\cdot) \circ \hat{k}^{-1}(\cdot)$.
\end{proof}

\subsubsection{Backdoor Criterion (BC)}
\begin{lemma}
\label{lemma:bc}
BGMs $f$ and $\hat{f}$ that produce the same distribution $P_{\mathcal{D}}(\boldsymbol{X}, V)$ are equivalent if
\begin{enumerate}
    \item (BC) $U \indep \boldsymbol{X} | \boldsymbol{Z}$ and $\hat{U} \indep \boldsymbol{X} | \boldsymbol{Z}$.
    \item For every $\boldsymbol{x}: \nabla_{\boldsymbol{x}}|\det\boldsymbol{J}_{f^{-1}(\boldsymbol{x}, \cdot)}|$ and $\nabla_{\boldsymbol{x}}|\det\boldsymbol{J}_{\hat{f}(\boldsymbol{x}, \cdot)}|$ both exist.
    \item (Variability) $\forall u:$ Instances $\boldsymbol{z}_1, \ldots, \boldsymbol{z}_{d+1}$ exist such that $|\det \boldsymbol{M}_{\mathcal{D}}(u, \boldsymbol{z}_1, \ldots, \boldsymbol{z}_{d+1})| > 0,$ where
    \begin{equation}
    \boldsymbol{M}_{\mathcal{D}}(u, \boldsymbol{z}_1, \ldots, \boldsymbol{z}_{d+1}) \triangleq    
    \begin{bmatrix}
    P_{\mathcal{D}}(u|\boldsymbol{z}_1) & \nabla_u P_{\mathcal{D}}(u|\boldsymbol{z}_1)\\
    \vdots                & \vdots                 \\
    P_{\mathcal{D}}(u|\boldsymbol{z}_{d+1}) & \nabla_u P_{\mathcal{D}}(u|\boldsymbol{z}_{d+1})\\
    \end{bmatrix}
    \end{equation}
\end{enumerate}
\end{lemma}
\begin{proof}
Define $g(\boldsymbol{x}, \cdot) \triangleq   f^{-1}(\boldsymbol{x}, \cdot) \circ \hat{f}(\boldsymbol{x}, \cdot)$.
Using the change of variable formula, we get
\begin{align}
&\forall \boldsymbol{x}, \hat{u}, \boldsymbol{z}: P_{\hat{U}|\boldsymbol{Z},\boldsymbol{X}}(\hat{u}|\boldsymbol{z},\boldsymbol{x}) = 
P_{U|\boldsymbol{Z},\boldsymbol{X}}\Big(g(\boldsymbol{x}, \hat{u})|\boldsymbol{z},\boldsymbol{x}\Big)|\det \boldsymbol{J}_{g(\boldsymbol{x}, \cdot)}|\\
\text{(BC)} \Rightarrow &\forall \boldsymbol{x},\hat{u},\boldsymbol{z}: P_{\hat{U}|\boldsymbol{Z}(\hat{u}|\boldsymbol{z})} = 
P_{U|\boldsymbol{Z}}\Big(g(\boldsymbol{x}, \hat{u})|\boldsymbol{z}\Big)|\det \boldsymbol{J}_{g(\boldsymbol{x}, \cdot)}|
\label{eqn:39}
\end{align}
Using chain rule of derivatives, we know that
$|\det\boldsymbol{J}_{g(\boldsymbol{x}, \cdot)}| = |\det\boldsymbol{J}_{f^{-1}(\boldsymbol{x}, \cdot)}||\det\boldsymbol{J}_{\hat{f}(\boldsymbol{x}, \cdot)}|$
which is differentiable with respect to $\boldsymbol{x}$ according to condition 2.
By differentiating \Cref{eqn:39} with respect to the $i^{\text{th}}$ element in $\boldsymbol{x}$ $(x_i)$ we get
\begin{align}
\nabla_u P_{U|\boldsymbol{Z}}\Big(g(\boldsymbol{x}, \cdot)|\boldsymbol{z}\Big)
\begin{bmatrix}
\frac{g_1(\boldsymbol{x}, \hat{u})}{\partial x_i}\\
\vdots\\
\frac{g_d(\boldsymbol{x}, \hat{u})}{\partial x_i}
\end{bmatrix}
|\det \boldsymbol{J}_{g(\boldsymbol{x}, \cdot)}|
+
P_{U|\boldsymbol{Z}}\Big(g(\boldsymbol{x}, \hat{u})|\boldsymbol{z}\Big)\frac{\partial |\det \boldsymbol{J}_{g(\boldsymbol{x}, \cdot)}|}{\partial x_i}
&=0\\
\Rightarrow
\begin{bmatrix}
\nabla_u P_{U|\boldsymbol{Z}}\Big(g(\boldsymbol{x}, \cdot)|\boldsymbol{z}\Big) & P_{U|\boldsymbol{Z}}\Big(g(\boldsymbol{x}, \hat{u})|\boldsymbol{z}\Big)
\end{bmatrix}
\begin{bmatrix}
\frac{g_1(\boldsymbol{x}, \hat{u})}{\partial x_i}\\
\vdots\\
\frac{g_d(\boldsymbol{x}, \hat{u})}{\partial x_i}\\
\frac{\partial |\det \boldsymbol{J}_{g(\boldsymbol{x}, \cdot)}|}{\partial x_i}
\end{bmatrix}
&= 0
\end{align}

Stacking the equations for $\boldsymbol{z}_1, \ldots, \boldsymbol{z}_{d+1}$ we get
\begin{equation}
\boldsymbol{M}_{\mathcal{D}}(u, \boldsymbol{z}_1, \ldots, \boldsymbol{z}_{d+1})
\begin{bmatrix}
\frac{\partial |\det \boldsymbol{J}_{g(\boldsymbol{x}, \cdot)}|}{\partial x_i}\\
\frac{g_1(\boldsymbol{x}, \hat{u})}{\partial x_i}\\
\vdots\\
\frac{g_d(\boldsymbol{x}, \hat{u})}{\partial x_i}
\end{bmatrix}
=
\begin{bmatrix}
0\\ \vdots \\0
\end{bmatrix}
\end{equation}

Since the square matrix is full-rank due to variability condition, all elements of the vector must be zero.
This means that $g(\boldsymbol{x}, \hat{u})$ does not depend on $x_i$.
Iterating the same argument for all $i \in \{1, \ldots, d\}$ we conclude that $g(\boldsymbol{x}, \hat{u})$ does not depend on $\boldsymbol{x}$ which concludes the proof.
\end{proof}

\subsection{\Cref{thm:intervention}}
\label{app:proof:thm:intervention}
BGM $f$ is counterfactually identifiable given $P_{\boldsymbol{X}, V}$ if
\begin{enumerate}
    \item (Markovian) $U \indep \boldsymbol{X}$.
    \item for all $\boldsymbol{x}$, $f(\boldsymbol{x}, \cdot)$ is either a strictly increasing or a strictly decreasing function.
\end{enumerate}
\begin{proof}
Let $\mathbb{F}$ be the class of BGMs that satisfy theorem's conditions. Consider any two BGMs $\hat{f}, f \in \mathbb{F}$ that produce the same distribution $P_{\mathcal{D}}(\boldsymbol{X}, V)$.
Using \Cref{lemma:intervention} we conclude their equivalence.
As a result, they produce the same counterfactuals (\Cref{prop:counterfactual-equivalence}), which establishes identifiability according to its definition in \S\ref{sec:prelim}.
\end{proof}

\subsubsection{Independence assumption is not sufficient by itself for counterfactual identifiability}
\label{app:subsub:counter-example}

In this section, we use a simple example to demonstrate that the independence assumption alone (without the monotonicity assumption) is not enough for BGM identification.
This example is taken from \citet[Sec.~3]{non-ident}
Consider the following two simple BGMs $f$ and $\hat{f}$:
\begin{equation}
    X \sim Bernoulli(0.5),\; \; U \sim Unif(0, 1), \; X \indep U, \; f = 
    \begin{cases} 
    U, &X = 1\\
    U-1, &X = 0          
    \end{cases}
    , \; \hat{f} =
    \begin{cases} 
    U, &X = 1\\
    -U, &X = 0          
    \end{cases}
\label{eqn:simple_scm}
\end{equation}

Note that $f$ and $\hat{f}$ generate the same distribution $P_{\mathcal{D}}(\boldsymbol{X}, V)$, and they satisfy the first (independence) constraint.
However, they give different answers to counterfactual queries. 
Consider the following counterfactual query: $V_{1}|X=0, V=v$.
$f$ and $\hat{f}$ give $v+1$ and $-v$ as answers, respectively.

\subsection{\Cref{thm:IV}}
\label{app:proof:thm:IV}

For $\boldsymbol{X} \in \mathbb{X} \triangleq \{\boldsymbol{x}_1, \ldots, \boldsymbol{x}_n\}$ and $\boldsymbol{I} \in \mathbb{I} \triangleq \{\boldsymbol{i}_1, \ldots, \boldsymbol{i}_n\}$, BGM $f$ is counterfactually identifiable given $P_{\boldsymbol{X}, V, \boldsymbol{I}}$ if
\begin{enumerate}
    \item (IV) $\boldsymbol{I} \indep U$.
    \item for all $\boldsymbol{x} \in \mathbb{X}$, $f(\boldsymbol{x}, \cdot)$ and $f^{-1}(\boldsymbol{x}, \cdot)$ are either strictly increasing or strictly decreasing, and two times differentiable.
    \item $P(\boldsymbol{i}, \boldsymbol{x}, \cdot)$ is differentiable for every $\boldsymbol{i}\in \mathbb{I}, \boldsymbol{x}\in \mathbb{X}$.
    \item (Positivity) $\forall u, \boldsymbol{x} \in \mathbb{X}: P_{U, \boldsymbol{X}}(u, \boldsymbol{x}) > 0$.
    \item (Variability) $\forall u: |\det \boldsymbol{M}(u, \mathbb{I})| \ge c > 0$ , where 
    \begin{equation*}
    \boldsymbol{M}(u, \mathbb{I}) \triangleq    
    \begin{bsmallmatrix}
    P(\boldsymbol{x}_1|u,\boldsymbol{i}_1) & \ldots & P(\boldsymbol{x}_n|u,\boldsymbol{i}_1) \\
    \vdots                   & \ddots & \vdots                   \\
    P(\boldsymbol{x}_1|u,\boldsymbol{i}_n) & \ldots & P(\boldsymbol{x}_n|u,\boldsymbol{i}_n) \\
    \end{bsmallmatrix}
    \end{equation*}
\end{enumerate}

\begin{proof}
Let $\mathbb{F}$ be the class of BGMs that satisfy theorem's conditions. Consider any two BGMs $\hat{f}, f \in \mathbb{F}$ that produce the same distribution $P_{\mathcal{D}}(\boldsymbol{X}, V)$.
Using \Cref{lemma:IV} we conclude their equivalence.
As a result, they produce the same counterfactuals (\Cref{prop:counterfactual-equivalence}), which establishes identifiability according to its definition in \S\ref{sec:prelim}.
\end{proof}

\subsection{\Cref{thm:BC}}
\label{app:proof:thm:BC}
BGM $f$ is counterfactually identifiable given $P_{\boldsymbol{X}, V, \boldsymbol{Z}}$ if
\begin{enumerate}
    \item (BC) $U \indep \boldsymbol{X} | \boldsymbol{Z}$.
    \item  $\forall \boldsymbol{x}: \nabla_{\boldsymbol{x}}|\det\boldsymbol{J}_{f(\boldsymbol{x}, \cdot)}|$ and $\nabla_{\boldsymbol{x}}|\det\boldsymbol{J}_{f^{-1}(\boldsymbol{x}, \cdot)}|$ exist.
    \item (Variability) $\forall u:$ Instances $\boldsymbol{z}_1, \ldots, \boldsymbol{z}_{d+1}$ exist such that $|\det \boldsymbol{M}(u, \boldsymbol{z}_1, \ldots, \boldsymbol{z}_{d+1})| > 0,$ where
    \begin{equation*}
    \small{\boldsymbol{M}(u, \boldsymbol{z}_1, \ldots, \boldsymbol{z}_{d+1})} \triangleq    
    \begin{bsmallmatrix}
    P(u|\boldsymbol{z}_1) & \nabla_u P(u|\boldsymbol{z}_1)\\
    \vdots                & \vdots                 \\
    P(u|\boldsymbol{z}_{d+1}) & \nabla_u P(u|\boldsymbol{z}_{d+1})\\
    \end{bsmallmatrix}
    \end{equation*}
\end{enumerate}

\begin{proof}
Let $\mathbb{F}$ be the class of BGMs that satisfy theorem's conditions. Consider any two BGMs $\hat{f}, f \in \mathbb{F}$ that produce the same distribution $P_{\mathcal{D}}(\boldsymbol{X}, V)$.
Using \Cref{lemma:bc} we conclude their equivalence.
As a result, they produce the same counterfactuals (\Cref{prop:counterfactual-equivalence}), which establishes identifiability according to its definition in \S\ref{sec:prelim}.
\end{proof}

\section{Directed Graphical Models}
\label{app:CI}
We can determine (Conditional) independencies of a joint distribution from its directed graphical model using d-seperation. 

\subsection{Backdoor Criterion (BC)}

\begin{figure}
    \centering
    \begin{subfigure}{0.33\linewidth}
        \includegraphics[width=\linewidth]{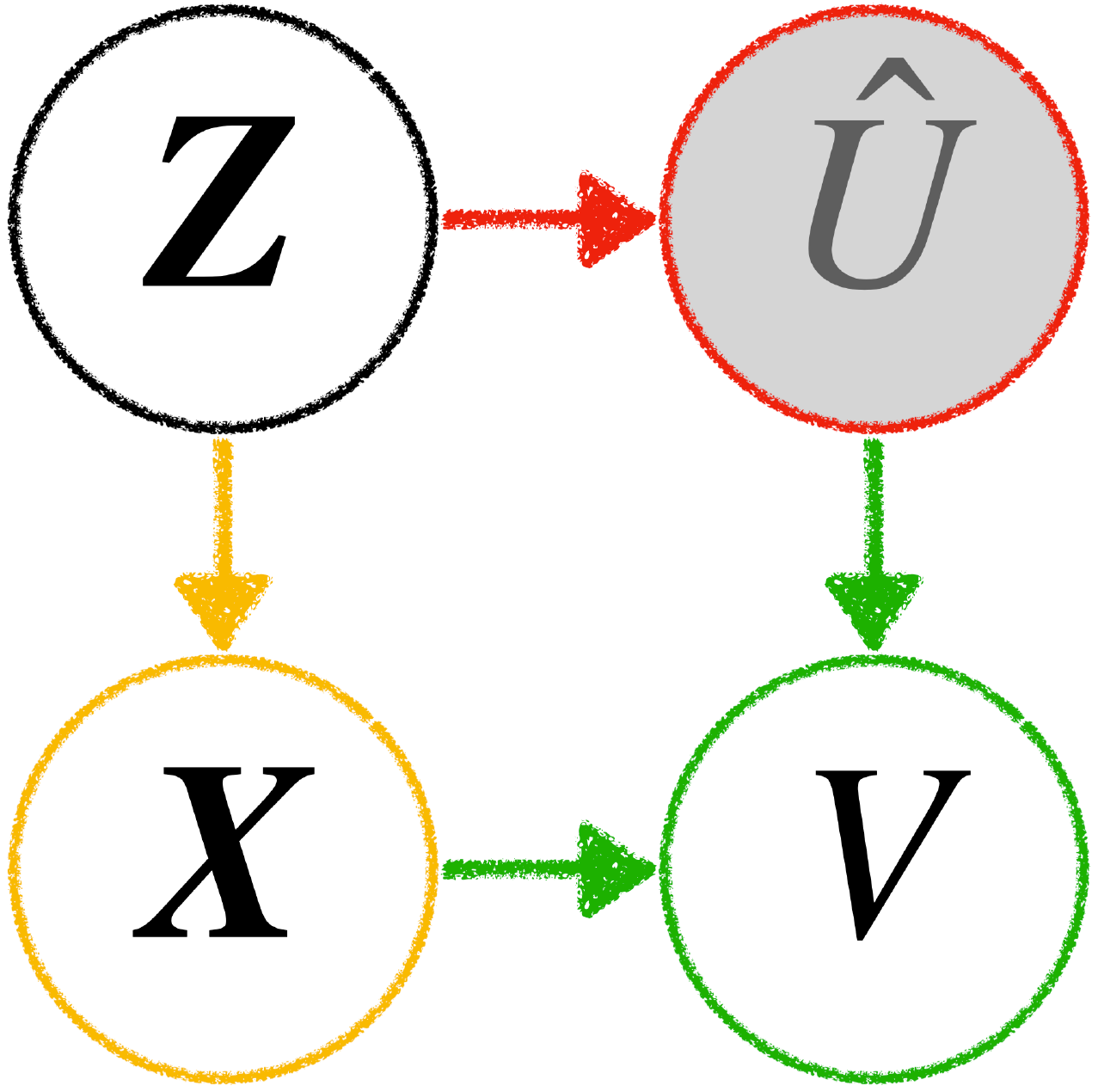}
        \caption{}
    \end{subfigure}
    \begin{subfigure}{0.33\linewidth}
        \centering
        \includegraphics[width=\linewidth]{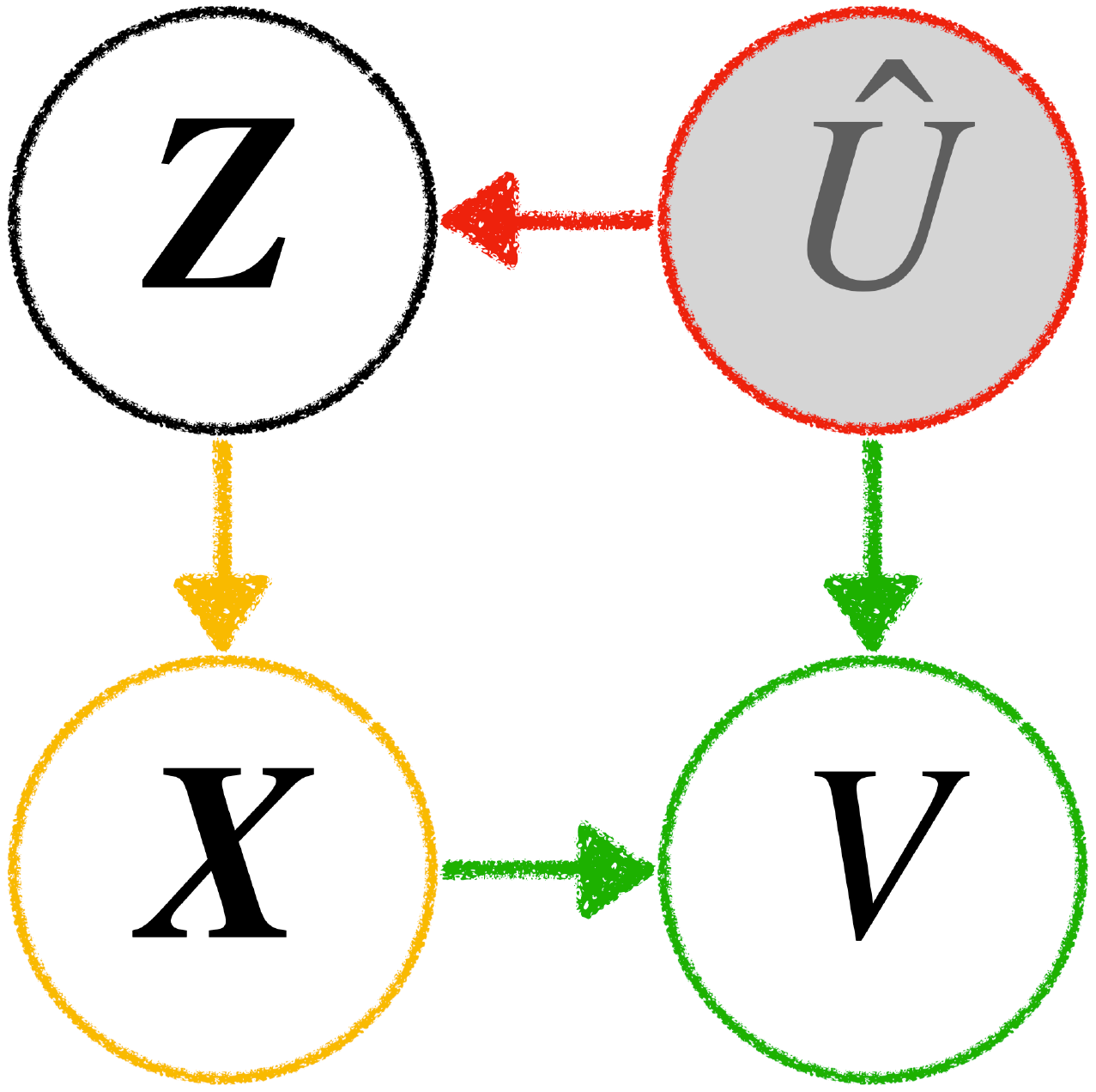}
        \caption{}
    \end{subfigure}
    \begin{subfigure}{0.33\linewidth}
        \centering
        \includegraphics[width=\linewidth]{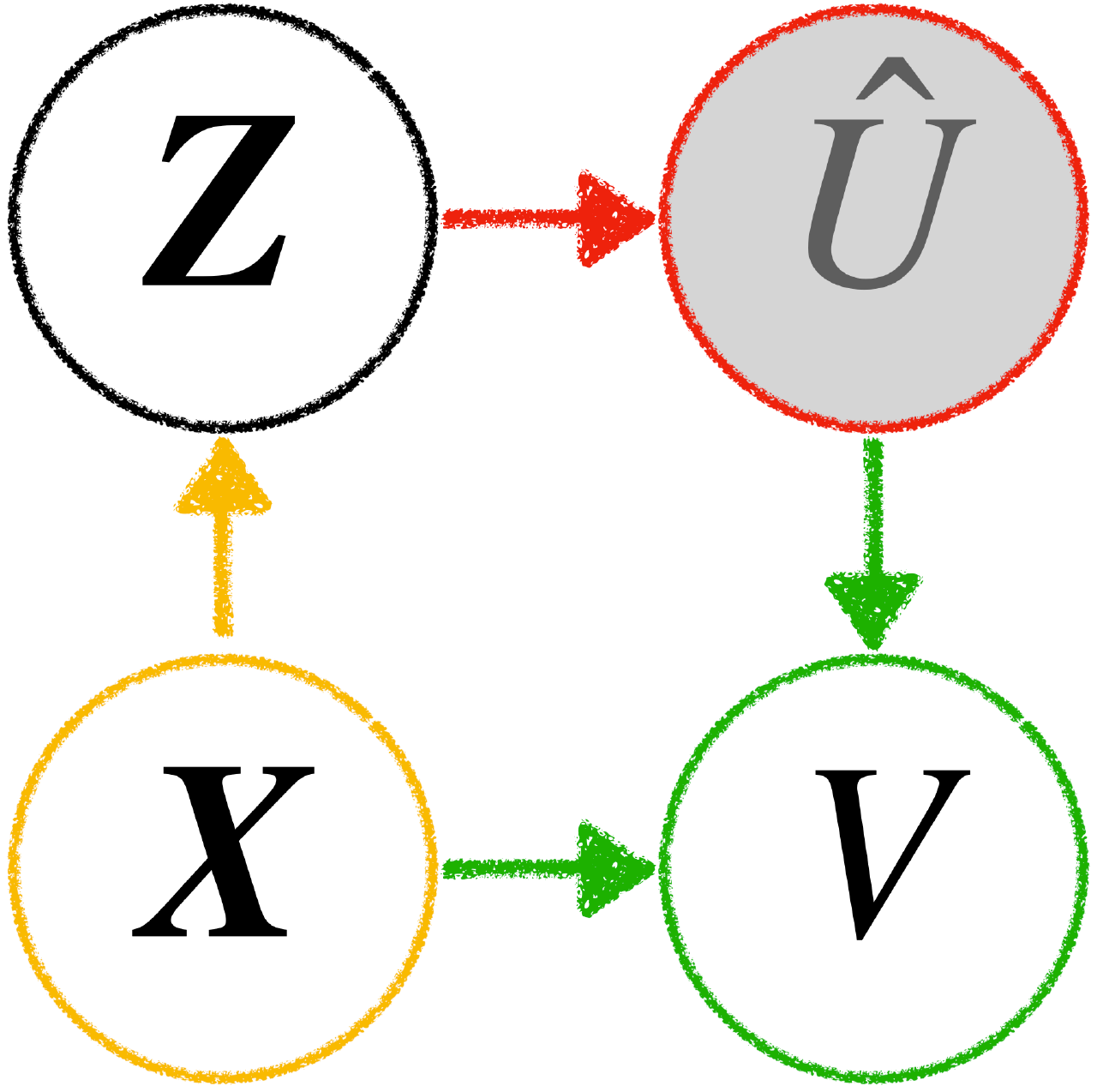}
        \caption{}
    \end{subfigure}
    \caption{Markovian equivalence class of directed graphical models for the backdoor criterion (BC) case}
    \label{app:fig:equivalence}
\end{figure}

The distributional requirement we have in this case (\S\ref{sub:BC}) is that $\boldsymbol{X} \indep U | \boldsymbol{Z}$.
The structured generative network we build for this case (\Cref{subfig:CNF-BC}) resembles the structure of the directed graphical model shown in \Cref{subfig:PGM:CNF-BC}.
As a result, it inherits all the conditional independence properties read off the graphical model using the $d$-separation test~\cite{pearl1988probabilistic,koller2009probabilistic}.
There are two paths between $U$ and $\boldsymbol{X}$, $\boldsymbol{X} \leftarrow \boldsymbol{Z} \rightarrow U$ which is blocked when we condition on $\boldsymbol{Z}$, and $\boldsymbol{X} \rightarrow V \leftarrow U$ which is blocked due to the v-structure at $V$, thus $\boldsymbol{X} \indep U | \boldsymbol{Z}$.

\textbf{Alternative viable graphical models}: 
It is worth emphasizing that the directed graphical model used for constructing the structured generative network is purely a statistical object as opposed to a causal DAG, and its goal is solely to equip the structured generative network with the corresponding distributional constraint in each case.
As a result, all directed graphical models in the Markovian equivalence class (directed graphical models that encode the same conditional independencies) of \Cref{subfig:PGM:CNF-BC} are valid and can be used for constructing alternative structured generative network.

To create the Markovian equivalence class, we should keep the graph's skeleton fixed, and flip the edges without creating new or removing existing v-structures.
\Cref{app:fig:equivalence} shows the three possible options, each of which we can use to construct a valid structured generative network.

\subsection{Instrumental Variable (IV)}
The structured generative network in this case is depicted in \Cref{subfig:CNF-IV}.
It is designed to follow the directed graphical model in \Cref{subfig:PGM:CNF-IV} in which the two paths between $\boldsymbol{I}$ and $U$, $\boldsymbol{I} \rightarrow \boldsymbol{X} \leftarrow U$ and $\boldsymbol{I} \rightarrow \boldsymbol{X} \rightarrow V \leftarrow U$ are blocked by open v-structures (unconditioned $X$ and $\boldsymbol{V}$, respectively) which implies independence of $\boldsymbol{I}$ and $V$
As a result, the distribution produced by the structured generative network is guaranteed to satisfy the distributional constraint $U \indep V$ (the first condition in \Cref{thm:IV}), by construction.

In this setting, the graphical model shown in \Cref{subfig:PGM:CNF-IV} is the only member of its Markovian equivalence class, as flipping any edges would change the set of (conditional) independencies.

\section{Normalizing Flows (NF)}
\label{app:NF}
Normalizing Flows (NF) are a class of generative models with tractable distributions where both sampling and density estimation are efficient and exact. 
They model the data ($V$) as a transformation ($T$) of some noise variable $(U)$ sampled from a simple base distribution ($P_U$), e.g., Gaussian distribution, where $T$ is a diffeomorphism.\footnote{A differentiable transform with a differentiable inverse.}
This allows for the density of $V$ to be obtained via a change of variables:
\begin{equation}
\label{eqn:change-of-var}
P_V(v) = P_U\Big(T^{-1}(v)\Big)|\det \boldsymbol{J}_{T^{-1}}(v)|
\end{equation}
The transform $T$ can be tractably optimized to fit the observed distribution of $V$.
Designing expressive transformation families with efficient inverse and Jacobin has thus been subject to research~\cite{kingma2018glow,chen2019residual,meng2022butterflyflow}.
This idea can be easily extended for modeling conditional distributions~\cite{trippe2018conditional,winkler2019learning,lu2020structured}, e.g., $P_{V|\boldsymbol{X}}$ with conditional normalizing flows (CNF), by parameterizing the transform $T$ as a function of the condition ($\boldsymbol{x}$).
Refer to \citet{kobyzev2020normalizing,papamakarios2021normalizing} for an extensive survey of NFs.

\section{Experiments}
\label{app:exp}

\textbf{Implementation Details}:
We build all CGMs using NFs with linear rational splines~\cite{dolatabadi2020invertible} and train them with likelihood maximization using their implementation in Pyro~\cite{bingham2018pyro}.
All splines we use have 16 bins for mapping $(-3, +3)$ to $(-3, +3)$.
We use affine transforms at input and output layers to calibrate the range.
Condition networks are all MLPs with two hidden layers, each with 64 units.
We use batch size of $2^{20}$ and run all experiments using A100 GPUs.
We train all models using the default implementation of Adam~\cite{kingma2015adam} in Pytorch~\cite{paszke2019pytorch}.

\textbf{Empirical Relaxation of Theoretical Assumptions}:
Linear rational splines, although differentiable, are not necessarily two times differentiable.
Two times differentiability is required in the IV case (\S\ref{sub:IV}) by the second assumption of \Cref{thm:IV}.
Furthermore, the condition network we use to condition the spline parameters based on $\boldsymbol{x}$ uses ReLU activation function~\cite{agarap2018deep} which implies non-differentiability of our CGMs with respect to $x$, which is required in the IV case (\S\ref{sub:BC}) by the second assumption of \Cref{thm:BC}.
We can satisfy both of these assumptions, e.g., by using quadratic splines~\cite{durkan2019neural} or GElU activations~\cite{hendrycks2016gaussian}.
However, good performance in \S\ref{sec:eval} suggests that these technical assumptions might not be tight, and can be relaxed.
We leave their relaxation to future work.

\subsection{Counterfactual Ellipse Generation}
\label{app:ellipse}
We use the following SCM for generation of the ellipse dataset:
\begin{align}
Z &\coloneqq \epsilon_z, \; \epsilon_z \sim \text{Unif}(-0.5, 0.5)\\
X &\coloneqq (1.44254843z + 0.59701923 + \epsilon_x) ~\%~ (2\pi), \; \epsilon_x \sim \text{Normal}(0, 1)\\
U_0 &\coloneqq e^{1.64985274z + 0.2656131} + \epsilon_{u_0}, \; \epsilon_{u_0} \sim \text{Beta}(1, 1)\\
U_1 &\coloneqq U_0(1 + \epsilon_{u_1}e^{1.61323358z - 0.18070237}), \; \epsilon_{u_1} \sim \text{Exponential}(1)\\
V_0 &\coloneqq U_0\Big(2 + \sin(X)\Big)\\
V_1 &\coloneqq U_1\Big(2 + \cos(X)\Big)\\
\end{align}
We use a sequence of three Spline transforms with coupling for all schemes.

\subsubsection{Failure in the Markovian Case}
\label{app:ellipse:failure}
In the ellipse generation taks, both $U$ and $V$ are two dimensional.
To empirically evaluate whether or not we can lean multi-dimensional BGMs in the Markovian case, we generated a second dataset by randomly shuffling $X$ in the previous dataset.
We trained a single CGM (similar to the Markovian case in \S\ref{sec:learning}), and used it for counterfactual estimation where it failed ($\text{MAPE} = 607$).

\subsection{Video Streaming Simulation}
\label{app:video}

We got the simulator and the ABR algorithms' implementations from \citet{causalsim}.
Appendix.~D in this work explains the details.
We use a single conditional spline for every CGM.

\subsubsection{Simultaneous Exploitation of Instrumental Variable (IV) and Backdoor Criterion (BC)}
\label{app:video:IV+BC}
\begin{figure}
    \centering
    \begin{subfigure}{0.54\linewidth}
        \includegraphics[width=\linewidth]{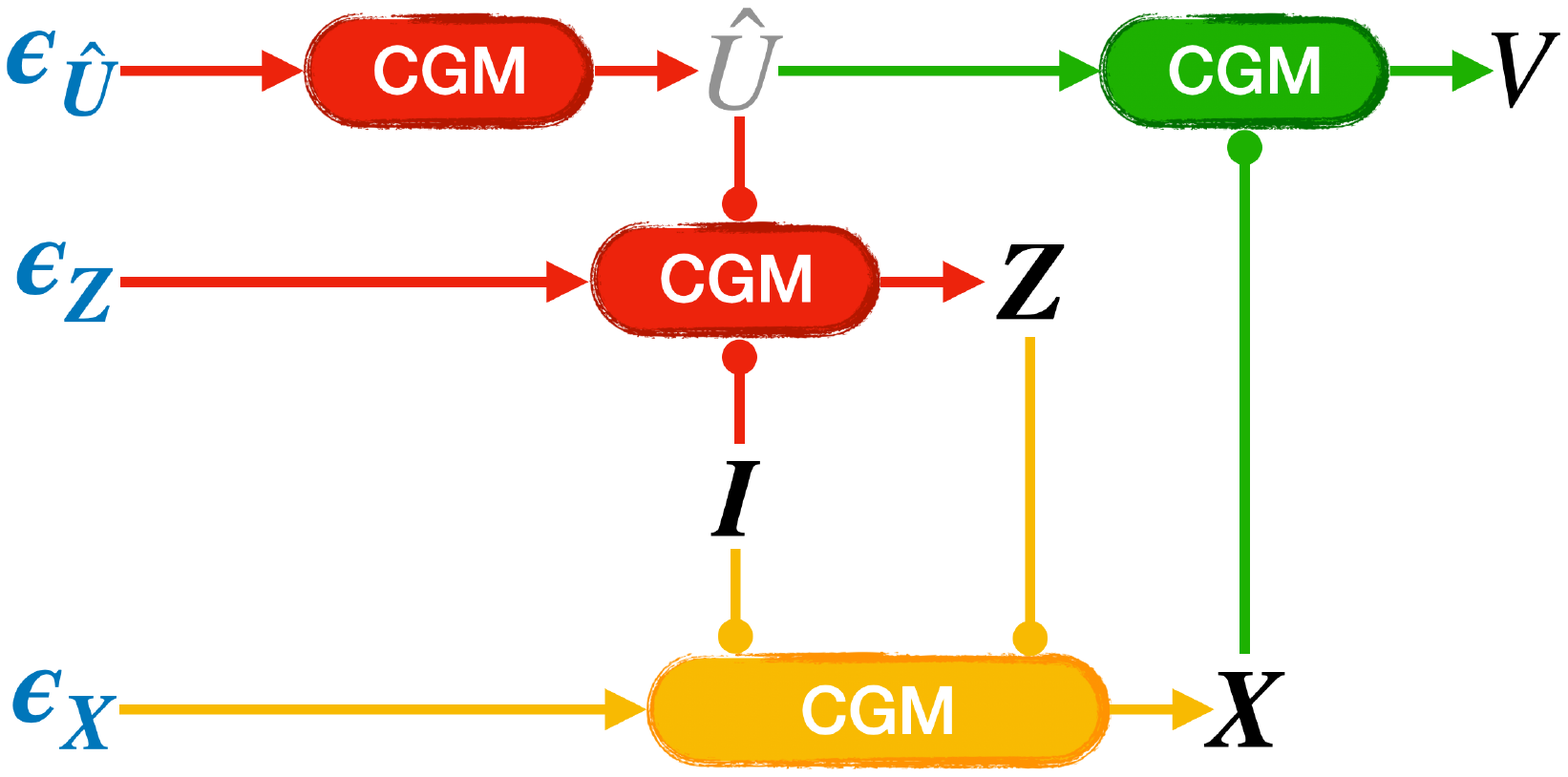}
        \vspace{-1cm}
        \caption{Structured Generative Network}
        \label{app:subfig:CNF-IV+BC}
    \end{subfigure}
    \begin{subfigure}{0.45\linewidth}
        \centering
        \includegraphics[width=\linewidth]{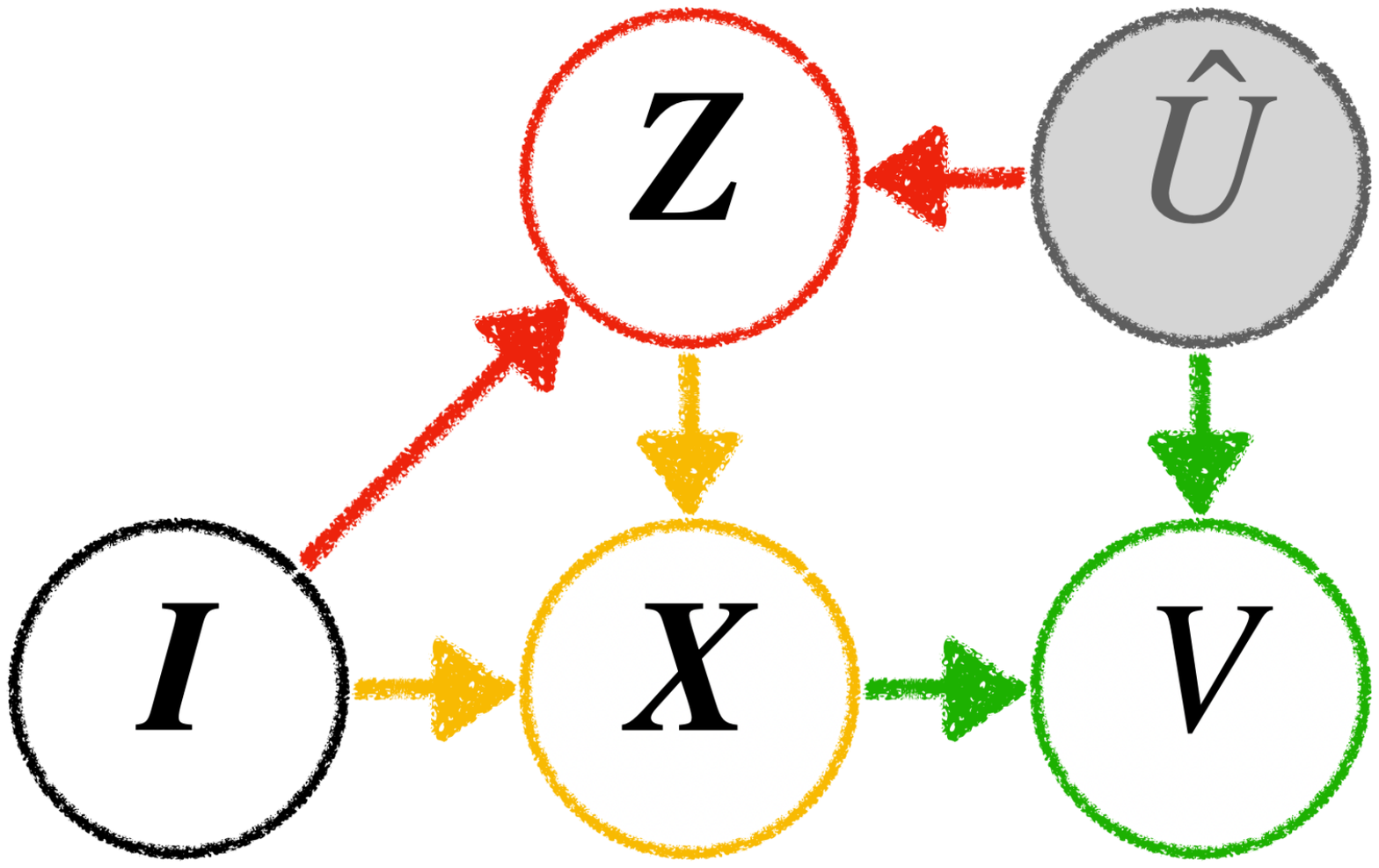}
        \caption{Directed Graphical Model}
        \label{app:subfig:PGM:CNF-IV+BC}
    \end{subfigure}
    \caption{Simultaneous Exploitation of Instrumental Variable (IV) and Backdoor Criterion (BC)}
    \label{app:fig:IV+BC}
\end{figure}

\Cref{app:subfig:PGM:CNF-IV+BC} depicts the directed graphical model we use to represent the necessary (conditional) independencies in this case.
In this graphical model, all paths between $\boldsymbol{I}$ and $\hat{U}$ are blocked so $\boldsymbol{I} \indep \hat{U}$ which is the distributional objective of the IV case (\S\ref{sub:IV}).
Furthermore, conditioning on $(\boldsymbol{Z}, \boldsymbol{I})$ blocks all paths between $\boldsymbol{X}$ and $U$ hence $\boldsymbol{X} \indep U | (\boldsymbol{I}, \boldsymbol{Z})$.
This is the distributional objective in the BC case (\S\ref{sub:BC}), where the pair $(\boldsymbol{I}, \boldsymbol{Z})$ satisfies the backdoor criterion.

\end{document}